\documentclass[journal,twoside,web]{ieeecolor}
\usepackage{tmi}
\usepackage{cite}
\usepackage{amsmath,amssymb,amsfonts}
\usepackage{algorithmic}
\usepackage{graphicx}
\usepackage{textcomp}
\usepackage{soul}

\usepackage{colortbl}
\usepackage{bbold}
\usepackage{bm}
\usepackage[ruled, vlined]{algorithm2e}
\usepackage{multirow}
\usepackage{pifont}
\usepackage{hhline}
\usepackage{array}
\usepackage{tabularx}
\usepackage{hhline}
\usepackage{svg}

\DeclareMathOperator*{\argmin}{\arg\!\min}
\newcommand{\cmark}{\ding{51}}%

\def\BibTeX{{\rm B\kern-.05em{\sc i\kern-.025em b}\kern-.08em
    T\kern-.1667em\lower.7ex\hbox{E}\kern-.125emX}}
\begin{document}
\title{COSST: Multi-organ Segmentation with Partially Labeled Datasets Using Comprehensive Supervisions and Self-training}  
\author{Han Liu, Zhoubing Xu, Riqiang Gao, Hao Li, Jianing Wang, Guillaume Chabin, Ipek Oguz, and Sasa Grbic
\thanks{Manuscript submitted on XX April 2023. This work was done during the research internship at Siemens Healthineers, Princeton, NJ USA.}
\thanks{Han Liu and Ipek Oguz are with the Department of Computer Science, Vanderbilt University, TN 37235 USA (e-mail: han.liu@vanderbilt.edu; ipek.oguz@vanderbilt.edu).}
\thanks{Hao Li is with the Department of Electrical and Computer Engineering, Vanderbilt University, Nashville, TN 37235 USA. (e-mail: hao.li.1@vanderbilt.edu).}
\thanks{Zhoubing Xu, Riqiang Gao, Jianing Wang, Guillaume Chabin and Sasa Grbic are with Siemens Healthineers, Princeton, NJ 08540 USA (e-mail: zhoubing.xu@siemens-healthineers.com; riqiang.gao@siemens-healthineers.com; jianing.wang@siemens-healthineers.com; guillaume.chabin@siemens-healthineers.com; sasa.grbic@siemens-healthineers.com).}}

\maketitle

\begin{abstract}
Deep learning models have demonstrated remarkable success in multi-organ segmentation but typically require large-scale datasets with all organs of interest annotated. However, medical image datasets are often low in sample size and only partially labeled, i.e., only a subset of organs are annotated. Therefore, it is crucial to investigate how to learn a unified model on the available partially labeled datasets to leverage their synergistic potential. In this paper, we systematically investigate the partial-label segmentation problem with theoretical and empirical analyses on the prior techniques. We revisit the problem from a perspective of partial label supervision signals and identify two signals derived from ground truth and one from pseudo labels. We propose a novel two-stage framework termed COSST, which effectively and efficiently integrates \underline{co}mprehensive \underline{s}upervision signals with \underline{s}elf-\underline{t}raining. Concretely, we first train an initial unified model using two ground truth-based signals and then iteratively incorporate the pseudo label signal to the initial model using self-training. To mitigate performance degradation caused by unreliable pseudo labels, we assess the reliability of pseudo labels via outlier detection in latent space and exclude the most unreliable pseudo labels from each self-training iteration. Extensive experiments are conducted on one public and three private partial-label segmentation tasks over 12 CT datasets. Experimental results show that our proposed COSST achieves significant improvement over the baseline method, i.e., individual networks trained on each partially labeled dataset. Compared to the state-of-the-art partial-label segmentation methods, COSST demonstrates consistent superior performance on various segmentation tasks and with different training data sizes. 




\end{abstract}


\section{Introduction}
\label{sec:introduction}
\IEEEPARstart{M}{ulti-organ} segmentation for computed tomography (CT) scans is a fundamental yet challenging task in medical imaging \cite{landman2015miccai,kavur2021chaos,ma2022fast,ma2021abdomenct}. It plays a crucial role in a variety of biomedical tasks. For example, in radiotherapy treatment planning, accurate delineation of organs at risk is clinically imperative and critical to guarantee a safe and effective treatment \cite{sahiner2019deep}. It also enables extraction of quantitative information such as organ shape and size for biomedical research\cite{schoppe2020deep}. Typically, delineation of critical organs needs to be performed manually by radiation oncologists but this process is highly tedious, time consuming, and prone to intra- and inter-observer variations. It is thus favorable to have automatic and accurate algorithms to perform the multi-organ segmentation task. 

\begin{figure}[t]
    \centerline{\includegraphics[width=1\columnwidth]{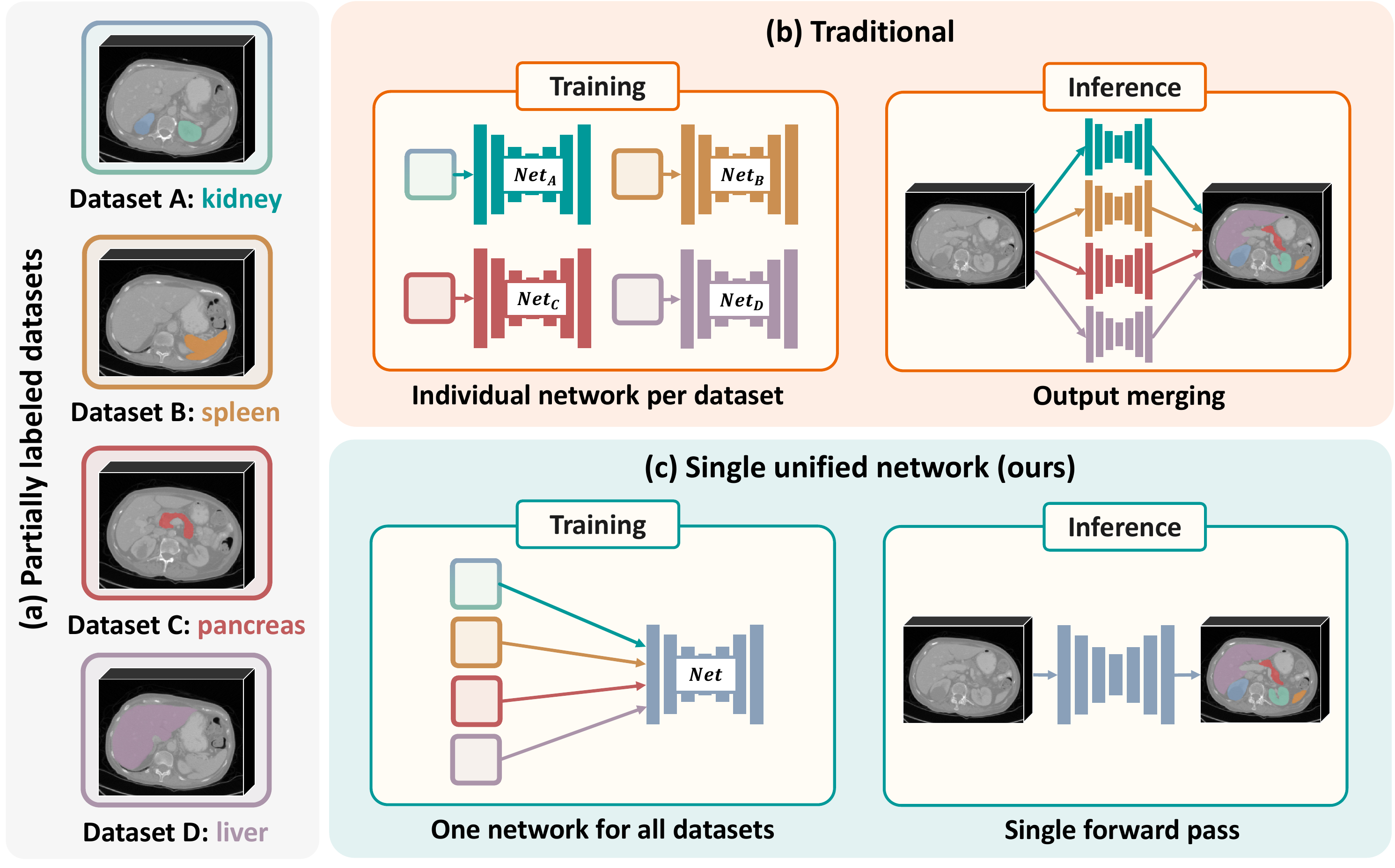}}
    \caption{(a) Practically, the datasets collected from different sites may only contain the annotations of a single or a few organs depending on the particular clinical purpose, and thus these datasets are consider partially labeled. The goal of partial-label segmentation is to segment all annotated structures by training on the partially labeled datasets. (b) The traditional method aims to train individual networks on each dataset and perform output merging during inference. (c) Our proposed method aims to learn a single unified network from all datasets and thus can be better scaled to a large number of partially labeled datasets.}
    \label{fig1}
\end{figure}


To date, deep learning models have achieved state-of-the-art performance on multi-organ segmentation tasks \cite{isensee2021nnu,gibson2018automatic,wang2019abdominal,tang2021high,lee2023d}, but they typically require all organs of interest to be annotated. However, due to the costly and laborious labeling process, it is extremely difficult to obtain a large-scale fully-annotated dataset, especially in the medical domain. In practice, medical image datasets are usually annotated by only a single or few organs depending on the particular clinical purpose at different institutes. Given a multi-organ segmentation task, these datasets are considered as \textit{partially labeled datasets}, which can be integrated to segment a full coverage of organs of interest (Fig. \ref{fig1}). Hence, it is highly desirable to develop an effective integration strategy to leverage the synergistic potential of the partially labeled datasets.

 

An intuitive strategy is to train individual models on each partially labeled dataset. Afterwards, the segmentation result of all requested organs can be obtained by ensembling the outputs from individual networks. An alternative strategy is to train a single unified model with multiple partially labeled datasets, where the organs of interest can be segmented simultaneously. In comparison, the latter strategy yields three clear advantages. First, based on the demonstrated benefits of larger training datasets for deep learning models\cite{deng2009imagenet}, a unified model trained on the union of all partially labeled datasets, is anticipated to outperform the individual models trained on each partially labeled dataset. Second, during deployment, using a single unified model can lead to faster inference speeds and reduced storage requirements. Lastly, it does not require extra post-processing steps to address conflicting voxel predictions (a voxel being predicted as different classes), a challenge that may arise when using multiple models.

Consequently, increasing efforts have been made to the unified models over the past few years\cite{zhou2019prior,fang2020multi,shi2021marginal,liu2022universal,huang2020multi,feng2021ms,zhang2021dodnet,dmitriev2019learning}. For instance, some studies proposed to address the partial-label segmentation problem from a perspective of network designs \cite{dmitriev2019learning,zhang2021dodnet}, e.g., conditioned networks, where the segmentation task from each partially labeled dataset is encoded as a task-aware prior to guide the model to segment on-demand organs. Other studies have attempted to tackle this problem from a perspective of using class adaptive losses \cite{fang2020multi,shi2021marginal} or pseudo label learning \cite{liu2022universal,huang2020multi,feng2021ms}. Nevertheless, there lacks a systematic understanding of the partial-label segmentation problem and an in-depth analysis of the existing techniques. Besides, we observe that most existing methods are developed and validated using singly-annotated datasets \cite{zhou2019prior,fang2020multi,shi2021marginal,huang2020multi,feng2021ms,zhang2021dodnet,dmitriev2019learning}, i.e., one annotated organ per dataset, whereas in practice a partially labeled dataset may have multiple annotated organs. Hence, additional validation of the well-established methods on multi-organ partially labeled datasets is needed.


In this study, we revisit the partial-label segmentation problem from a novel perspective, i.e., supervision signals, and ask ourselves two questions: (1) how many types of partial label supervision signals exist? and (2) how to leverage these signals for training? To answer the first question, we perform in-depth analyses on all mainstream partial-label segmentation approaches and identify \underline{three} distinct types of supervision signals, including two supervision signals derived from ground truth annotations and one from pseudo labels (see more in Sec. \ref{sec:II}). To answer the second question, we propose a novel two-stage framework named COSST, which effectively and efficiently integrates \underline{\textbf{co}}mprehensive \underline{\textbf{s}}upervision signals with \underline{\textbf{s}}elf-\underline{\textbf{t}}raining. Specifically, we propose to firstly train an initial unified model using two ground-truth based signals and then iteratively incorporate the pseudo label signal to the initial model using self-training. To mitigate the potential performance degradation caused by poor pseudo labels, we assess the reliability of pseudo labels and exclude the training data with detected unreliable pseudo labels at each self-training iteration. The pseudo label assessment approach is inspired by a unique property of partially labeled datasets: for each organ, ground truth annotations are available in at least one of the partially labeled datasets. Given a distribution of ground truth labels, the quality of a pseudo label can be assessed via outlier detection in latent space. In summary, the key contributions of our work are as follows:

\begin{itemize}
\item We systematically investigate the partial-label segmentation problem with both theoretical and empirical analyses on the prior techniques, identifying three distinct types of supervision signals.
\item We propose a novel two-stage framework for learning from partially labeled datasets, where comprehensive supervision signals are integrated effectively and efficiently via self-training.
\item Based on a unique property of partially labeled datasets, we design a novel pseudo label assessment and filtering strategy via outlier detection in latent space, further optimizing the usage of pseudo labels.
\item We perform extensive experiments on one public and three private partial label segmentation tasks over 12 CT datasets, demonstrating the effectiveness of COSST and our pseudo label assessment strategy.
\end{itemize}

\begin{figure}[t]
    \centerline{\includegraphics[width=1\columnwidth]{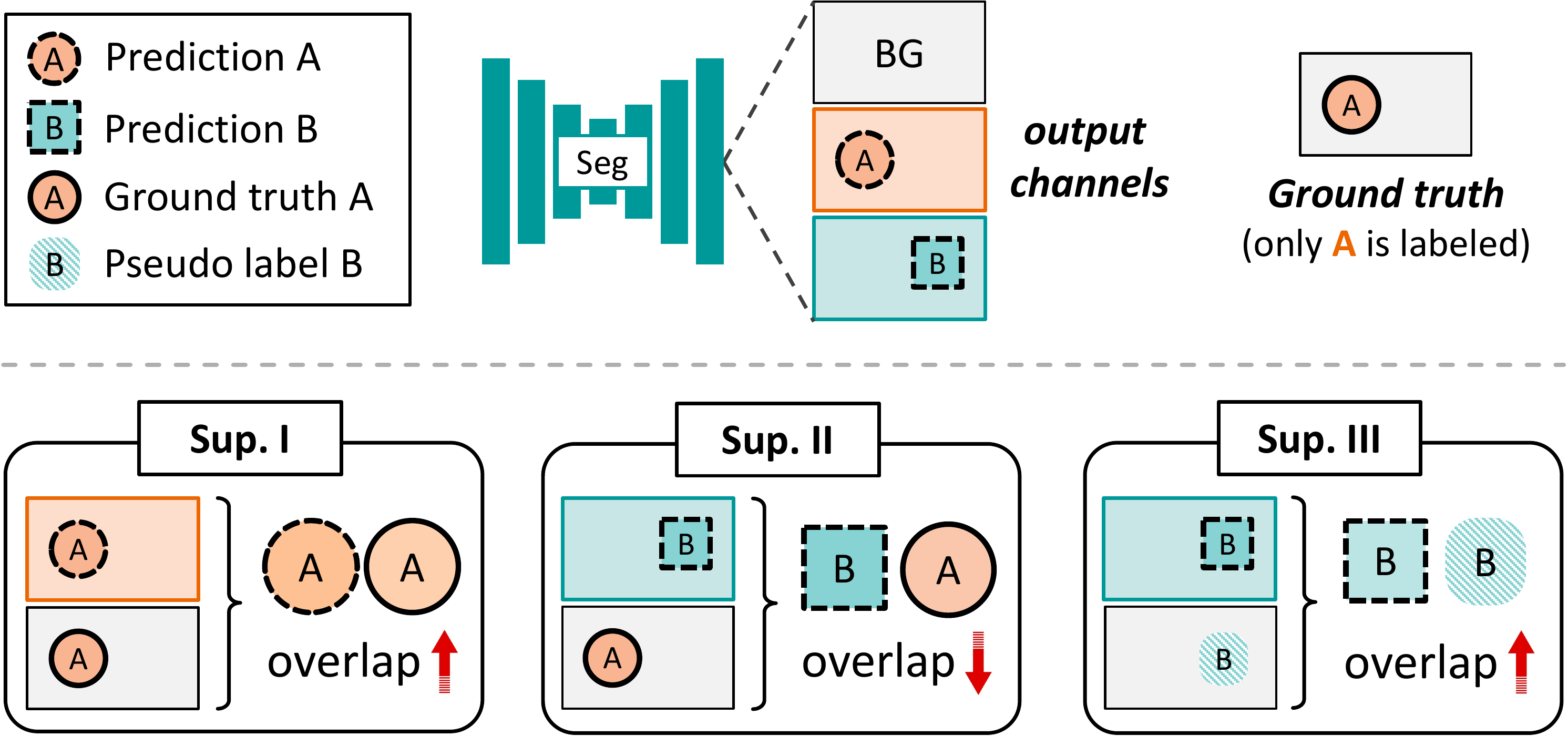}}
    \caption{Illustration of three types of partial label supervision signals. Given an input image where only organ A is labeled, Sup. I aims to maximize the overlap between output A and ground truth A. Sup. II aims to minimize the overlap between output B and ground truth A (due to mutual exclusiveness among organs). Sup. III aims to maximize the overlap between output B and pseudo label B.}
    \label{fig2}
\end{figure}

\section{Supervision Signals in Partial-Label Segmentation}
\label{sec:II}
With a systematic analysis of the existing partial-label segmentation approaches, we summarize that there are primarily three types of supervision signals, denoted as Sup. I, II and III. In Fig. \ref{fig2}, we illustrate these supervision signals with a toy example. Imagine there are two partially labeled datasets: Dataset $\mathbb{A}$ (labeled with organ A) and Dataset $\mathbb{B}$ (labeled with organ B). For a multi-class segmentation network (typically with a softmax function), there are three output channels corresponding to background (BG), A and B. Now, consider an image from \underline{Dataset $\mathbb{A}$} being passed to the network. 

\textbf{Sup. I} aims to \textbf{\textit{maximize}} the overlap between the prediction of the labeled organ (organ A) and the corresponding ground truth. This signal utilizes the available annotations to supervise labeled organs as in a standard segmentation task.

\textbf{Sup. II} aims to \textbf{\textit{minimize}} the overlap between the prediction of the unlabeled organ (organ B) and the ground truth of the labeled organ (organ A). This is inspired by the fact that different organs must be mutually exclusive \cite{shi2021marginal}. In other words, each foreground voxel must be classified as either A or B in our example. The mutual exclusiveness can thus be used as a constraint to regularize the predictions of unlabeled organs based on the available labeled organs.

\textbf{Sup. III} aims to \textbf{\textit{maximize}} the overlap between the prediction and the pseudo label for the unlabeled organ (organ B). Compared to Sup. II, where the prediction of the unlabeled organ is constrained to where it \textit{cannot} overlap, Sup. III imposes a stronger supervision by guiding the prediction to where it \textit{should} overlap, i.e., pseudo labels. Note that pseudo labels can be easily generated by the models trained on individual partially labeled dataset.

\textbf{Discussion} Sup. I is applied to the labeled organs whereas Sup. II and III are applied to the unlabeled organs. Besides, we note that Sup. I and II are derived from ground truth annotations, whereas Sup. III is derived from pseudo labels. Compared to Sup. III, which can be noisy due to unreliable pseudo labels, Sup. I and II are noise-free throughout the training process. This observation motivates us to separate supervision signals into different training stages in our COSST. 


\begin{table}[t]
\centering
\caption{Comparison of types of supervision signals used in COSST and the existing partial label segmentation methods.}
\label{tab1}
\begin{tabular}{|c|c|c|c|} 
\hline
Method & Sup. I & Sup. II & Sup. III \\ 
\hline
TAL \cite{fang2020multi}& \cmark &  &  \\ 
\hline
ME \cite{shi2021marginal}& \cmark & \cmark &  \\ 
\hline
PLT \cite{liu2022universal}& \cmark &  & \cmark \\ 
\hline
Co-training \cite{huang2020multi}& \cmark &  & \cmark \\ 
\hline
DoDNet \cite{zhang2021dodnet}& \cmark &  &  \\ 
\hline
MS-KD \cite{feng2021ms}&  &  & \cmark \\ 
\hline
COSST (ours) & \cmark & \cmark & \cmark \\
\hline
\end{tabular}
\end{table}

\section{Related Works}
\label{sec:III}

\subsection{Partially Labeled Medical Image Segmentation}
In the past few years, substantial efforts have been devoted to explore partially labeled image segmentation. A straightforward strategy is to train individual networks on each partially labeled dataset, but suffers from several drawbacks including: (1) less training data for each individual network, (2) longer inference time, and (3) more complexity for post-processing. 

Recent studies have been focused on training a single unified model with multiple partially labeled datasets. Zhou \textit{et al.} \cite{zhou2019prior} proposed Prior-aware Neural Network (PaNN) by firstly estimating anatomical priors of organ sizes based on a fully labeled dataset, and then regularizing the organ size distributions on the partially labeled datasets. However, this approach requires access to at least one fully annotated dataset and thus cannot generalize well. Some studies have attempted to design adaptive loss functions that can be directly applied to partially labeled data\cite{gonzalez2018multi,fang2020multi,shi2021marginal}. Fang \textit{et al.} \cite{fang2020multi} presented a target adaptive loss (TAL) by treating the unlabeled organs as background. In addition, Shi \textit{et al.} \cite{shi2021marginal} proposed a marginal and exclusive loss by imposing an additional exclusive constraint for the unlabeled organs. Since these adaptive loss functions are designed to learn exclusively from the labeled organs, the inherent limitation of these methods lies in their oversight of the substantial potential offered by pseudo labeling for the unlabeled organs in the partially labeled datasets.

To unleash the potential of pseudo labeling, several works have explored to incorporate pseudo label learning to learn from both the labeled and unlabeled organs\cite{liu2022universal,huang2020multi,petit2021iterative,feng2021ms}. Liu \textit{et al.} \cite{liu2022universal} proposed to first train individual models on each partially labeled dataset and generate pseudo labels for the unlabeled organs. Then a pseudo multi-organ dataset, consisting of both ground truth and pseudo labels, was used for supervised training. Huang \textit{et al.} \cite{huang2020multi} developed a co-training framework based on cross-pseudo supervision\cite{chen2021semi}, where the prediction of unlabeled organs from one network is supervised by the weight-averaged output of the other network. However, these approaches overlook the assessment of pseudo label quality and thus may suffer from performance degradation caused by poor pseudo labels. Besides, Feng \textit{et al.} \cite{feng2021ms} presented a multi-teacher single-student knowledge distillation (MS-KD) framework by learning the soft pseudo labels generated by the teacher models pre-trained on each partially labeled dataset. Nevertheless, this method may not be applicable to a large number of partially labeled datasets due to the limit of GPU memory to load all teacher models.

The aforementioned methods rely on a standard multi-output channel network (typically with a softmax activation). Recently, conditioned networks have emerged as an effective alternative network architecture for partial-label segmentation \cite{dmitriev2019learning,zhang2021dodnet,deng2023omni,wu2022tgnet}, where a task-aware prior is used to guide the segmentation of the task-related organ. Dmitriev \textit{et al.} \cite{dmitriev2019learning} incorporated organ class information into the intermediate activation signal for training. Zhang \textit{et al.} \cite{zhang2021dodnet} presented a dynamic on-demand network (DoDNet) by using one-hot code as task-aware prior to generate weights for dynamic convolution filters. However, the conditioned networks are typically designed to segment one organ per forward pass and hence can be computationally inefficient.


Our proposed method is different from the existing techniques in three major aspects: (1) as shown in Tab. \ref{tab1}, compared to the existing approaches, COSST leverages more comprehensive supervision signals for training. (2) COSST employs self-training to better exploit the pseudo labels. (3) COSST explicitly considers the quality of pseudo labels during training, a crucial aspect that is often overlooked in other pseudo label-based approaches. 


\begin{figure*}[t]
\centerline{\includegraphics[width=1\textwidth]{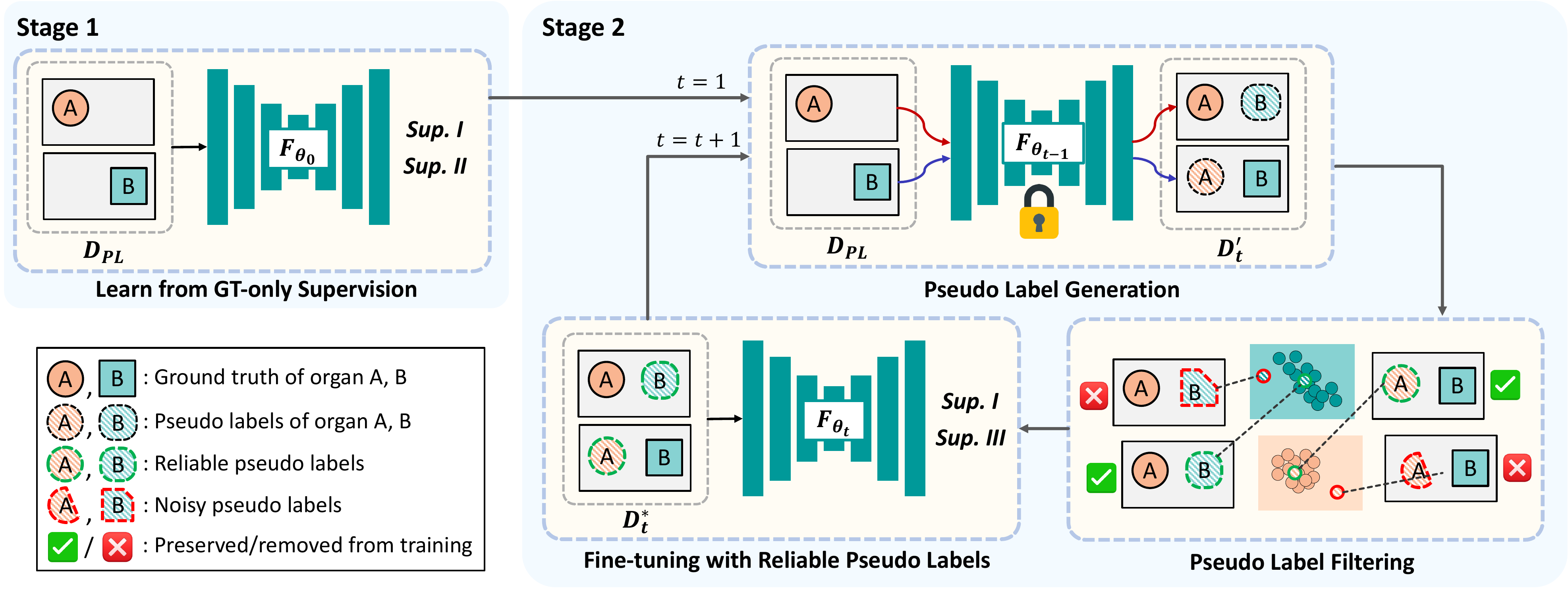}}
    \caption{An illustration of our proposed two-stage framework COSST. In stage 1, we train an initial unified model using only the ground truth-based supervision. In stage 2, we use self-training to iteratively incorporate the most updated pseudo label supervision to the initial model. In each iteration, we first create a pseudo multi-organ dataset by generating the pseudo labels for the unlabeled organs. Then we assess the quality of pseudo labels and perform image-level pseudo label filtering by removing the training data with unreliable pseudo labels from the pseudo multi-organ dataset. Lastly, we fine-tune the initial unified model on the filtered dataset. The self-training process is repeated iteratively until convergence.}
    \label{fig3}
\end{figure*}

\subsection{Self-training}
Self-training is an iterative process that aims to improve the model performance by exploiting the predictions of the model on unlabeled data, i.e., pseudo labels, and has been widely investigated in semi-supervised learning\cite{lee2013pseudo,yang2022st++,chaitanya2023local} and domain adaptation\cite{zou2019confidence,xie2022unsupervised,shin2022cosmos,dong2021unsupervised,liu2022enhancing}. In self-training, a model is first trained on a labeled dataset and then used to generate the pseudo labels on an unlabeled dataset. Lastly, the model is retrained on the human labels and pseudo labels jointly. As more unlabeled data is incorporated into the training process, the model performance is expected to improve. The improved model can in turn generate better pseudo labels and this process is repeated until convergence. Previous studies have shown that the performance of self-training can be further improved by using pseudo labels with higher confidence\cite{yang2022st++,shin2022cosmos,dong2021unsupervised}, where the unconfident images or pixels can be either de-prioritized\cite{yang2022st++} or removed from training\cite{shin2022cosmos,dong2021unsupervised}. This motivates us to develop a pseudo label filtering strategy for the self-training process in our COSST.

\section{Methods}
\label{sec:IV}
\subsection{Overview}
Let us consider $N$ partially labeled datasets ${\mathcal{D}}_{PL}=\{{\mathcal{D}}_{1}, {\mathcal{D}}_{2}, ..., {\mathcal{D}}_{N}\}$, which are annotated with $C_{1}, C_{2}, ..., C_{N}$ types of organs, respectively. We aim to learn a single unified segmentation model $\mathcal{F}$$_\theta$ from ${\mathcal{D}}_{PL}$ to segment $C_{PL}=\bigcup_{i}^{N}C_{i}$ organs. An illustration of our proposed COSST is shown in Fig. \ref{fig3}. Overall, COSST consists of two training stages: (i) learning from ground truth-based supervision and (ii) self-training with pseudo labels. To optimize pseudo label learning, we introduce a pseudo label assessment and filtering strategy to mitigate the potential performance degradation caused by noisy pseudo labels. Detailed descriptions are as follows.




\subsection{Learning from Ground Truth-based Supervision}
In stage 1, we aim to learn an initial unified model $\mathcal{F}_{\theta_{0}}$ using the supervision signals derived \textit{only} from the ground truth annotations: (i) Sup. I: for labeled organs, they can be supervised using the available annotations. (ii) Sup. II: for unlabeled organs, they can be supervised to not overlap the annotated regions. The major challenge is that there are always certain labels absent in partially labeled data, making the traditional segmentation losses inapplicable. To tackle this problem, we use adaptive loss functions as in \cite{fang2020multi,shi2021marginal}. Specifically, for labeled organs, we treat the unlabeled organs as background by merging the output channels of the original background channel and all unlabeled organs into a new background channel. The channels are merged by taking the sum of the probabilities. Given an input image $x$, the original model prediction $\tilde{y}=\mathcal{F}(x;\theta)$ is thus transformed to a new prediction $\tilde{y_{t}}$, which only has the output channels of the new background and the labeled organs, allowing regular segmentation losses to be directly applied. For unlabeled organs, we first create a binary mask $M$ by taking the union of all labeled organs. We then regularize all output channels of unlabeled organs by minimizing the overlap between the prediction on each channel and the binary mask. Let $y$ be the ground truth annotation, $C_{u}$ be the number of unlabeled organs, and $L$ be a standard segmentation loss, e.g., Dice loss. The overall learning objective of training the initial unified model can be expressed as:




\begin{equation}
\theta_{0}=\argmin_{\theta}{L(\tilde{y_{t}}, y)-\sum_{u\in{C_{u}}}{L(\tilde{y_{u}}, M)}}
\end{equation}

\subsection{Self-training with Pseudo Labels}
In the case of partially labeled data, pseudo labels of the unlabeled organs are inherently available and can be exploited without additional annotation efforts. However, we observe that pseudo labels are either overlooked or not optimized in the existing partial-label segmentation approaches (Tab. \ref{tab1}). For example, in \cite{liu2022universal,huang2020multi}, pseudo labels are generated by the individual networks trained on each partially labeled dataset. As demonstrated later in our experiments, our initial unified model $\mathcal{F}_{\theta_{0}}$ obtained in the first training stage can already outperform the individual networks, suggesting that better pseudo labels can be used. Indeed, the quality of pseudo labels is highly dependent on the performance of the network used for pseudo label generation. This motivates us to use self-training to optimize the usage of pseudo labels. Specifically, at self-training iteration $t\in\{1,2,3...\}$, we use $\mathcal{F}_{\theta_{t-1}}$ to generate pseudo labels for unlabeled organs. We obtain the network prediction $\tilde{y}=\mathcal{F}(x; \theta_{t-1})$ as pseudo labels and then merge the pseudo labels of unlabeled organs to $y$, with the original ground truth of labeled organs retained. Note that during the label merging, ground truth has higher priority than pseudo labels when there are conflicting labels. As a result, we obtain a pseudo multi-organ dataset ${\mathcal{D}}^{'}_{t}$ where each training data is fully-annotated by the merged labels. Lastly, we obtain an improved model $\mathcal{F}_{\theta_{t}}$ by fine-tuning our initial model $\mathcal{F}_{\theta_{0}}$ on the pseudo multi-organ dataset ${\mathcal{D}}^{'}_{t}$. The learning objective of self-training at iteration $t$ can thus be expressed as:

\begin{equation}
\theta_{t}=\argmin_{\theta}{L(\tilde{y}, y^{'}_{t}; \theta_{0})}    
\end{equation}

\noindent{where} $y^{'}_{t}\in{\mathcal{D}}^{'}_{t}$. Note that, fine-tuning, which has shown its success in incorporating new pseudo labels to pre-trained network \cite{wang2016cost,kostic2022pseudo,song2020learning}, offers an efficient way to train $\mathcal{F}_{\theta_{0}}$ (trained by Sup. I and II) on the most updated pseudo labels (Sup. III). The self-training process is repeated iteratively until the model performance reaches plateaus on the validation set.


\subsection{Pseudo Label Assessment and Filtering}
The quality of pseudo labels plays a key role in self-training. As shown in previous studies, unreliable pseudo labels may lead to severe confirmation bias\cite{arazo2020pseudo} and potential performance degradation\cite{wang2021self,yang2022st++}. To address this problem, we develop a pseudo label assessment and filtering strategy to better exploit the pseudo labels during self-training, as illustrated in Fig. \ref{fig4}. Particularly, our assessment strategy is inspired by a unique property of partially labeled datasets: for each organ, ground truth annotations are available in at least one of the partially labeled datasets. Therefore, given the distribution of the available ground truth labels, the quality of a pseudo label can be assessed via outlier detection in latent space: if the pseudo label is a clear outlier deviating from the ground truth distribution, it is very likely to be a noisy label.

To this end, inspired by \cite{zhang2021prototypical}, we represent each organ (both labeled and unlabeled) in each training data as a feature vector by using the merged label $y^{'}$. Suppose the network input be $x\in\mathbb{R}^{ch\times{}h\times{}w\times{}d}$ with $ch$ channels, $h$ height, $w$ width and $d$ depth. A multi-class segmentation network $\mathcal{F}$ can be decomposed as (1) a dense feature extractor $\mathcal{F}$$_{feat}:\mathbb{R}^{ch\times{h}\times{w}\times{d}}\xrightarrow{}\mathbb{R}^{m\times{h}\times{w}\times{d}}$ and (2) a subsequent voxel-wise classifier $\mathcal{F}$$_{cls}:\mathbb{R}^{m\times{h}\times{w}\times{d}}\xrightarrow{}[0, 1]^{(1+C_{PL})\times{}h\times{}w\times{}d}$ that projects the $m$ dimensional features into class predictions. For the $i_{th}$ training data $x_{i}$, we first calculate the voxel-wise feature representation using the dense feature extractor $\mathcal{F}$$_{feat}$, where the feature representation of the $j_{th}$ voxel is expressed as $\mathcal{F}$$_{feat}(x_{i})^{j}$. For the $k_{th}$ organ, we obtain the organ-wise feature representation $z^{(i,k)}\in\mathbb{R}^{m}$ by computing the feature centroid for all voxels belonging to the mask $y^{'(k)}$:

\begin{figure}[t]
\centerline{\includegraphics[width=1\columnwidth]{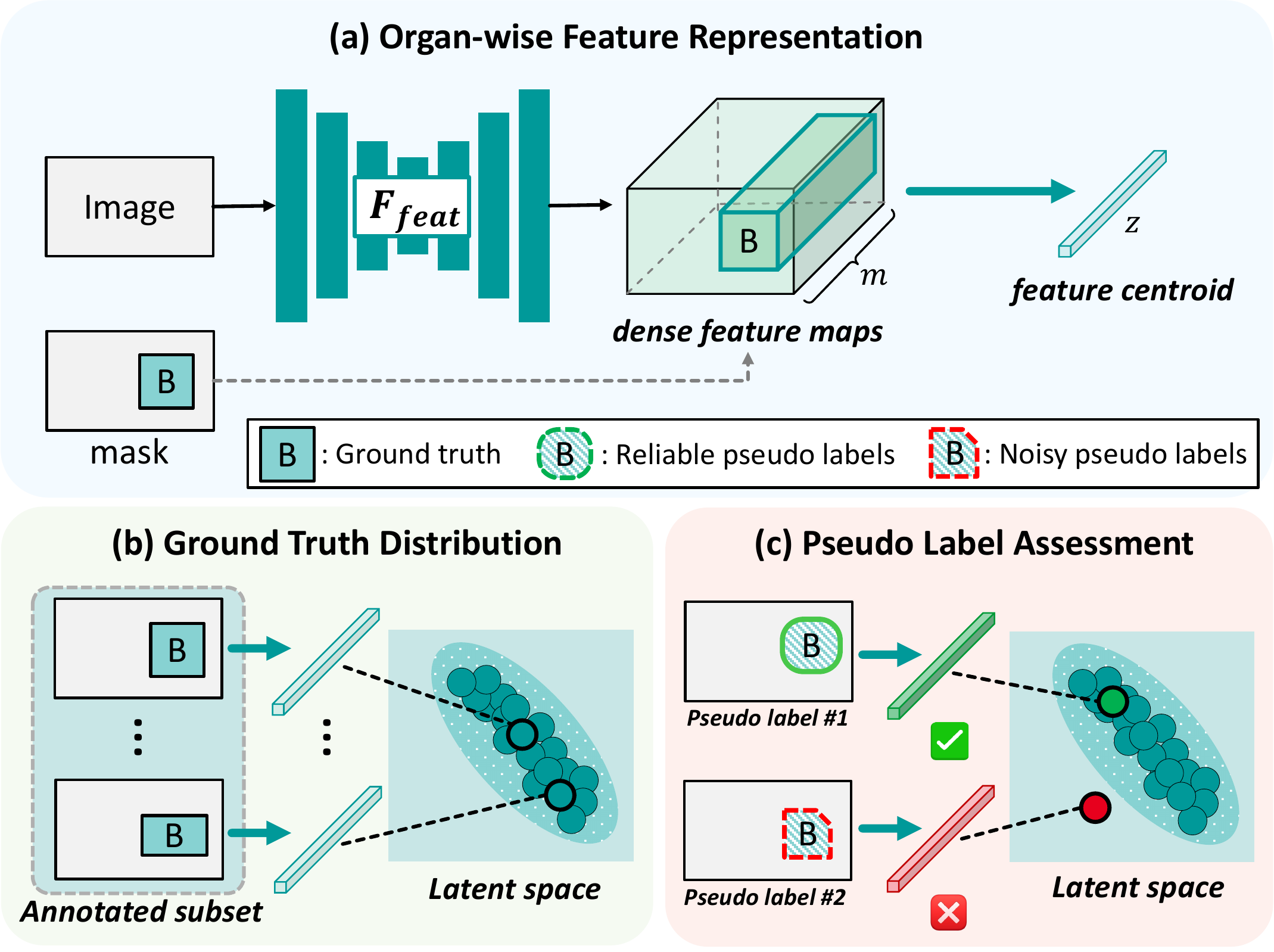}}
    \caption{(a) An illustration of how organ-wise feature representation $z$ is computed. The image is firstly passed to the feature extractor of the segmentation model to extract dense feature maps. The organ-wise feature representation is calculated as the feature centroid of the voxels within the organ of interest based on the mask. Note that this 'mask' can be either the ground truth annotation or the pseudo label. (b) Ground truth distribution for organ B (generated by the partially labeled datasets where organ B is annotated) in latent space. (c) The quality of pseudo labels is assessed via outlier detection in latent space.}
    \label{fig4}
\end{figure} 

\begin{equation}
\label{eq3}
    z^{(i,k)} = \frac {\sum_{j}\mathcal{F}_{feat}(x_{i})^{j} * \mathbb{1}(y^{'(j,k)}==1)} {\sum_{j}\mathbb{1}(y^{'(j,k)}==1)}
\end{equation}

\noindent{where} $\mathbb{1}$ is the indicator function. Besides, we use principal component analysis (PCA) to reduce the dimension of $z$ from $m$ to $2$, which is empirically more effective and computationally more efficient. Let $n$ be the number of training data with the $k_{th}$ organ annotated. The ground truth distribution for the $k_{th}$ organ can thus be expressed as: $
\bm{z^{k}}=\{z^{i,k},...,z^{i+n,k}\}$.

 Given the ground truth distribution $\bm{z^{k}}$, we aim to assess whether each pseudo label is an outlier. Prior studies show that the feature centroid of a distribution, or prototype, can be used to assess the similarity between the distribution and a query sample by measuring its distance to the prototype (typically by Euclidean distance) \cite{zhang2021prototypical, ruff2018deep}. However, our preliminary experiments show that this strategy is not effective for our tasks, possibly because representing the entire distribution as a single feature vector results in a loss of intricate intra-class relationship among samples. To address this problem, we propose to assess the reliability of pseudo labels using Mahalanobis distance, which considers the intra-class relationship by taking into account the covariance matrix. The Mahalanobis distance $d$ between the assessed pseudo label $\bar{z}^{k}$ and the ground truth distribution $\bm{z^{k}}$ can be expressed as: $d^{2}(\bar{z}^{k}, \mu, C)=(\bar{z}^{k}-\mu)^{T} \cdot{} C^{-1} \cdot{} (\bar{z}^{k}-\mu)$, where $\mu$ and $C$ represent the mean feature vector and covariance matrix of $\bm{z^{k}}$, respectively. To detect the outlier, we define a threshold $\tau$ for Mahalanobis distance and a pseudo label is considered unreliable if $d(\bar{z}^{k}, \mu, C) > \tau$.

The detected unreliable pseudo labels may cause performance degradation and thus need to be denoised or removed before training. Inspired by \cite{yang2022st++}, we propose to filter the unreliable pseudo labels on image-level. Specifically, we remove the training data with unreliable pseudo labels from the the pseudo multi-organ dataset ${\mathcal{D}}^{'}_{t}$, resulting in a filtered dataset ${\mathcal{D}}^{*}_{t}$. By incorporating the pseudo label filtering to self-training, we replace the overall learning objective of self-training at iteration $t$ from Eq. 2 to Eq. 4:
\begin{equation}
\theta_{t}=\argmin_{\theta}{L(\tilde{y}, y^{*}_{t}; \theta_{0})}    
\end{equation}

\noindent{where} $y^{*}_{t}\in{\mathcal{D}}^{*}_{t}$. As in classical self-training where the labeled and unlabeled data are optimized jointly, both the labeled (Sup. I) and unlabeled organs (Sup. III) in the preserved training data are optimized in Eq. 4. Note that COSST also mitigates the information loss caused by image-level filtering, i.e., the labeled organs in the filtered images are excluded from training, by fine-tuning on the initial unified model, where all labeled organs have been used as ground truth-based supervision signals in the first training stage. Lastly, the pseudo code for COSST is presented in Alg. \ref{alg1}.

\label{Sec:algorithm}
\SetKwInOut{Input}{input}
\SetKwInOut{Output}{output}
\begin{algorithm}[t]
\label{alg1}
  \DontPrintSemicolon
  \Input{Partially labeled datasets ${\mathcal{D}}_{PL}$,
  \\ hyperparameters: $\tau$}
  \Output{Parameters of the unified model $\theta$} 
  \textcolor{gray}{\small\tcp{Stage 1: learning from ground truth-based supervision}}
  Train an initial unified model on $\mathcal{D_{PL}}$ with Eq.1 $\xrightarrow{}$ $\theta_{0}$\;

  \textcolor{gray}{\small\tcp{Stage 2: self-training with reliable pseudo labels}}
  \Repeat{converge}{
    \For{$t=1:T$}{
        Generate pseudo multi-organ dataset with $\theta_{t-1}:$ ${\mathcal{D}}_{PL}\xrightarrow{}{\mathcal{D}}^{'}_{t}$\;
        Detect unreliable pseudo labels with Eq. 3. \;
        Image-level pseudo label filtering: ${\mathcal{D}}^{'}_{t}\xrightarrow{}{\mathcal{D}}^{*}_{t}$\;
        Fine-tune on $\mathcal{D}^{*}_{t}$ with Eq. 4: $\xrightarrow{}\theta_{t}$\;
        }} 
  \caption{Pseudocode of COSST}
\end{algorithm}


 

\section{Experiments}
\label{sec:V}

\subsection{Datasets}
\subsubsection{Public Datasets}
In this experiment, we use four public, partially labeled CT datasets for training, including (1) the task03 liver dataset from Medical Segmentation Decathlon (MSD)\cite{antonelli2022medical}, (2) the task09 spleen dataset from MSD, (3) the task07 pancreas dataset from MSD, and (4) the KiTS19 dataset \cite{heller2020state}. Following the experimental setting in \cite{shi2021marginal}, we merge the cancer label to the organ label for liver, pancreas and kidney datasets. In addition, we also manually divide the binary kidney masks into left and right kidneys through connected component analysis. In this task, our goal is to train a single unified segmentation model from the partially labeled datasets to segment all five organs, i.e., left kidney, right kidney, spleen, liver, and pancreas. This task will be referred to as \textbf{task1} in this paper. For evaluation, besides the held-out testing sets from the MSD and KiTS19 datasets, we perform additional evaluation on two external public CT datasets, i.e., BTCV \cite{landman2015miccai} and AMOS2022 \cite{ji2022amos}. This evaluation can be used to assess the model's generalizability to unseen CT datasets collected from different scanners and sites, which is common yet challenging in practice. To summarize, our testing set consists of a total number of 662 CT scans. Details of the involved datasets are shown in Tab. \ref{tab2}.

\begin{table*}
\centering
\caption{A summary description of the public datasets for \textbf{task1}. \# organs: the number of organs that are related to our task.}
\label{tab2}
\begin{tabular}{|c|c|c|c|c|} 
\hline
Dataset & \# train / valid / test & \# organs & annotated organs \\ 
\hline
KiTS19 & 84 / 21 / 105 & 2 & left kidney, right kidney \\ 
\hline
Spleen (MSD) & 16 / 4 / 21 & 1 & spleen \\ 
\hline
Panceas (MSD) & 112 / 28 / 140 & 1 & pancreas \\ 
\hline
Liver (MSD) & 52 / 13 / 66 & 1 & liver \\ 
\hline
BTCV & - / - / 30 & 5 & left kidney, right kidney, spleen, pancreas, liver, other structures \\ 
\hline
AMOS2022 & - / - / 300 & 5 & left kidney, right kidney, spleen, pancreas, liver, other structures \\ 
\hline
\end{tabular}
\end{table*}

\subsubsection{Private Datasets}
We also conduct experiments on our private CT datasets for three partial-label segmentation tasks. These tasks aim to segment more diverse sets of organs from three body regions, i.e., bowel, pelvic and eye regions, which are referred to as \textbf{task2}, \textbf{task3}, and \textbf{task4}, respectively. Specifically, each task consists of two partially labeled CT datasets and each dataset may contain multiple annotated organs. The datasets were acquired from regular radiotherapy planning routine and the organs were annotated by a team of experienced specialists with an internal annotation tool. For each organ, a detailed annotation protocol was set up based on Radiation Therapy Oncology Group (RTOG) guidelines. A quality assessment was performed for each annotated dataset before further use. In contrast to the existing studies where single-organ datasets are mostly used, our private partially labeled datasets are mostly annotated with multiple organs, leading to a rarely studied yet more challenging experimental setting. Details of our private datasets are shown in Tab. \ref{tab3}. 

We obtained the fully-annotated training sets of \underline{bowel datasets} by having specialists additionally annotate the unlabeled organs on the partially labeled datasets. Thus, the bowel datasets were also used to (1) compare the model trained with partially labeled datasets against fully-annotated datasets (upper bound) and (2) comprehensively evaluate the quality of pseudo labels as later shown in our ablation study. 

\begin{table*}
\centering
\caption{A summary description of our private datasets for \textbf{task2-4}. Each task aims to segment all annotated organs from two partially labeled datasets. L: Left. R: Right.}
\label{tab3}
\begin{tabular}{|c|c|c|c|c|c|c|} 
\hline
Task & Dataset & spacing (mm) & \# train / valid / test & \# organs & annotated organs \\ 
\hline
\multicolumn{1}{|c|}{\multirow{2}{*}{Task 2}} & Bowel 1 & $2\times2\times2$ & 104 / 41 / 63 & 2 & duodenum, small bowel \\ 
\cline{2-6}
\multicolumn{1}{|c|}{} & Bowel 2 & $2\times2\times2$ & 104 / 41 / 63 & 3 & large bowel, sigmoid, rectum \\ 
\hline
\multicolumn{1}{|c|}{\multirow{2}{*}{Task 3}} & Pelvic 1 & $2\times2\times2$ & 568 / 72 / 72 & 6 & bladder, prostate, rectum, femur (L), femur (R), seminal vesicle \\ 
\cline{2-6}
\multicolumn{1}{|c|}{} & Pelvic 2 & $2\times2\times2$ & 128 / 16 / 16 & 1 & uterus \\ 
\hline
\multicolumn{1}{|c|}{\multirow{2}{*}{Task 4}} & Eye 1 & $1\times1\times1$ & 124 / 62 / 63 & 3 & chiasm, optic nerve (L), optic nerve (R) \\ 
\cline{2-6}
\multicolumn{1}{|c|}{} & Eye 2 & $1\times1\times1$ & 125 / 62 / 63 & 4 & len (L), len (R), eyeball (L), eyeball (R) \\
\hline
\end{tabular}
\end{table*}

\subsection{Experiment Setup}
\subsubsection{Implementation Details}
For \textbf{task1}, we follow the experiment setting from \cite{shi2021marginal} and build upon the nnU-Net framework \cite{isensee2021nnu} for all compared methods. nnU-Net is selected due to its demonstrated superior performance across a spectrum of MICCAI segmentation challenges. The output layer of the network is activated by softmax and the number of output channels is set to $C_{PL}+1$. For preprocessing, all CT scans are adjusted to the RAI orientation, resampled to $1.5\times1.5\times3.0$ mm, clipped to $[-1024, 1024]$ Hounsfield Units (HU), and rescaled to $[0,1]$. We use the wide window for intensity clipping to ensure fair contrast for different types of organs such as soft tissues and bones. For all compared methods and the stage 1 of COSST, we use the same hyperparameters configured by nnU-Net for training, including a total number of 1000 training epochs, an initial learning rate of $0.01$, an optimizer of the stochastic gradient descent (SGD) algorithm with a Nesterov momentum (µ=0.99), and a learning rate scheduler of polynomial decay policy. The patch size is configured as $160\times192\times64$ by the nnU-Net. The model checkpoints with the best performance on the validation set are selected for final evaluation. For the stage 2 of COSST, i.e., fine-tuning with self-training, we set the maximum number of epochs as 200 and the initial learning rate as $0.0001$. The other hyperparameters and the model selection criteria are kept the same as the stage 1.

For \textbf{task2-4}, we adopt the classical 3D U-Net\cite{ronneberger2015u} as the backbone architecture and use the same preprocessing procedures and hyperparameters (learning rate, optimizer, and epochs) as in \textbf{task1}. During training, we randomly extract 3D patches with a fixed size of $128 \times{} 128 \times{} 128$ with the center being a foreground or background voxel using a ratio of $2:1$. To achieve optimal performance for all compared methods, we applied a variety of augmentation techniques on-the-fly including rotation, scaling, Gaussian blur, Gaussian noise, brightness, contrast, low resolution simulation and gamma correction. The sum of Dice loss and cross-entropy loss is used as the segmentation loss. During inference, we utilize the sliding window inference with a window step size equal to half of the patch size and the overlapping windows are merged using Gaussian weighting. For all segmentation tasks, we empirically set the threshold for Mahalanobis distance $\tau$ as $\chi^{2}(2,0.999)$, i.e., the 99.9\% quantile of the chi-squared distribution with a degree of freedom of 2. All experiments are implemented in PyTorch\cite{paszke2019pytorch} v1.10 and MONAI\cite{cardoso2022monai} v0.8 with a single NVIDIA V100 16 GB GPU.


\subsubsection{Evaluation Metrics}
Dice Similarity Coefficient (DSC), average symmetric surface distance (ASD), and 95th Hausdorff distance (HD95) are used to evaluate the segmentation performance. DSC computes the overlapping between the predicted mask and ground truth. ASD evaluates the quality of segmentation boundaries by computing the average of all distances between the predicted mask and the ground truth boundary. HD95 is the 95th percentile of the maximum distances between the boundary points in the prediction and the ground truth, which suppresses the impact of outlier voxels in the prediction. In our experiments, all the validation and testing sets (except for the external testing datasets for \textbf{task1}) are also partially labeled datasets. For these datasets, the average metrics for each organ are only computed based on the dataset where the organ is annotated. For the same reason, Wilcoxon signed-rank test used for statistical analysis is conducted on the metrics of individual organs for the held-out testing sets of MSD and KiTS19 in \textbf{task1} and all testing sets in \textbf{task2-4}, and on the average metrics of all organs for BTCV and AMOS2022 testing sets.

\newcolumntype{C}{ >{\centering\arraybackslash} m{0.54cm} }
\newcolumntype{D}{ >{\centering\arraybackslash} m{1.67cm} }

\begin{table*}
\centering
\caption{Performance comparison (DSC, \%, higher is better; HD95, mm, lower is better; ASD, mm, lower is better) of partially labeled segmentation methods for \textbf{task1} on the \underline{public} datasets. Red and blue denote the \textcolor{red}{best} and \textcolor{blue}{second best} scores. $\dagger$ and $*$ represent p-value <0.05 and <0.01 using Wilcoxon signed-rank test; otherwise, p-value >0.05. P: held-out testing sets of MSD and KiTS19. F1: BTCV. F2: AMOS2022.}
\label{tab4}
\begin{tabular}{|D|C|C|C|C|C|C|C|C|C|C|C|C|C|C|C|} 
\hline
\multirow{2}{*}{Methods} & \multicolumn{3}{c|}{Left kidney (P)} & \multicolumn{3}{c|}{Right kidney (P)} & \multicolumn{3}{c|}{Spleen (P)} & \multicolumn{3}{c|}{Pancreas (P)} & \multicolumn{3}{c|}{Liver (P)} \\ 
\cline{2-16}
 & DSC &HD95 & ASD & DSC &HD95 & ASD & DSC &HD95 & ASD & DSC &HD95 & ASD & DSC &HD95 & ASD \\ 
\hline
Multi-Nets & 96.11$^{*}$ & 6.20$^{*}$ & 1.27$^{*}$ & 96.20$^{*}$ & 3.13 & 0.58$^{*}$ & 91.59$^{*}$ & 26.15$^{*}$ & 3.72$^{*}$ & 77.21$^{\dagger}$ & 6.88$^{*}$ & 1.40 & 94.81$^{*}$ & 7.54$^{*}$ & 1.28\\ 
\hline
TAL\cite{fang2020multi} & 96.13$^{*}$ & 4.19$^{\dagger}$ & 0.81 & \textcolor{blue}{96.22}$^{*}$ & 4.05$^{\dagger}$ & 0.66 & 93.72 & 2.97 & 0.82 & 76.36$^{*}$ & 6.37 & 1.43$^{*}$ & 94.86 & \textcolor{blue}{6.10} & 1.31 \\ 
\hline
ME\cite{shi2021marginal} & \textcolor{blue}{96.16}$^{*}$ & \textcolor{red}{1.65}$^{*}$ & \textcolor{red}{0.44}$^{*}$ & 95.87$^{*}$ & 7.56$^{*}$ & 1.27$^{*}$ & \textcolor{blue}{94.09}$^{*}$ & \textcolor{blue}{2.38} & \textcolor{blue}{0.77}$^{\dagger}$ & 75.99$^{*}$ & 6.01$^{*}$ & \textcolor{red}{0.77}$^{*}$ & 94.70 & 7.45 & 1.90\\ 
\hline
PLT\cite{liu2022universal} & 95.97$^{*}$ & 3.29$^{*}$ &0.81$^{*}$ & 96.21$^{*}$ & \textcolor{blue}{2.46} & \textcolor{blue}{0.58}$^{*}$ & 93.41$^{*}$ & 34.85$^{*}$ & 3.97$^{*}$ & 77.69 & 6.82$^{\dagger}$ & 1.32 & \textcolor{blue}{95.24} & 6.13 & \textcolor{blue}{1.10}  \\ 
\hline
Co-train\cite{huang2020multi} & 95.89$^{*}$ & 4.99 & 0.91 & 96.35$^{*}$ & 4.79 & 0.70 & 93.09$^{*}$ & 42.06$^{*}$ & 5.24$^{*}$ & 77.50 & 6.22$^{\dagger}$ & 1.21 & \textcolor{red}{95.24} & 5.10$^{\dagger}$ & \textcolor{red}{0.96} \\ 
\hline
DoDNet\cite{zhang2021dodnet} & 95.33$^{*}$ & 9.23$^{*}$ & 1.21$^{*}$ & 95.12$^{*}$ & 16.13$^{*}$ & 3.43$^{*}$ & 93.75 & 15.44$^{\dagger}$ & 2.23 & \textcolor{red}{79.09}$^{*}$ & \textcolor{blue}{5.84} & \textcolor{blue}{1.17}$^{*}$ & 93.65$^{*}$ & 11.26$^{*}$ & 2.03$^{*}$ \\ 
\hline
COSST (ours) & \textcolor{red}{96.55} & \textcolor{blue}{2.76} & \textcolor{blue}{0.51} & \textcolor{red}{96.48} & \textcolor{red}{2.03} & \textcolor{red}{0.42} & \textcolor{red}{94.40} & \textcolor{red}{1.45} & \textcolor{red}{0.54} & \textcolor{blue}{77.98} & \textcolor{red}{5.82} & 1.29 & 95.22 & \textcolor{red}{5.78} & 1.18 \\ 
\hline
\end{tabular}

\begin{tabular}{|D|C|C|C|C|C|C|C|C|C|C|C|C|C|C|C|} 
\hline\hline
\multirow{2}{*}{Methods} & \multicolumn{3}{c|}{Left kidney (F1)} & \multicolumn{3}{c|}{Right kidney (F1)} & \multicolumn{3}{c|}{Spleen (F1)} & \multicolumn{3}{c|}{Pancreas (F1)} & \multicolumn{3}{c|}{Liver (F1)} \\ 
\cline{2-16}
 & DSC &HD95 & ASD & DSC &HD95 & ASD & DSC &HD95 & ASD & DSC &HD95 & ASD & DSC &HD95 & ASD \\ 
\hline
Multi-Nets & 86.34 & 21.61 & 4.94 & 84.58 & 16.08 & 5.82 & 87.72 & 29.36 & 6.21 & 73.11 & 22.06 & 4.14 & 94.80 & 7.19 & 1.18\\ 
\hline
TAL\cite{fang2020multi} & 87.03 & 12.74 & 2.48 & 83.51 & 15.52 & 5.42 & 90.02 & \textcolor{blue}{12.15} & \textcolor{blue}{2.28} & 73.86 & 7.77 & 1.76 & \textcolor{red}{95.99} & 8.01 & 1.05 \\ 
\hline
ME\cite{shi2021marginal} & 85.62 & \textcolor{blue}{9.52} & \textcolor{red}{1.70} & 84.50 & \textcolor{blue}{10.20} & \textcolor{blue}{2.60} & \textcolor{blue}{90.58} & 14.25 & 2.91 & 74.87 & 21.77 & 3.65 & 95.96 & 10.15 & 1.60\\ 
\hline
PLT\cite{liu2022universal} & \textcolor{blue}{87.29} & 10.55 & \textcolor{blue}{2.42} & 84.13 & 15.46 & 5.20 & 88.00 & 55.87 & 8.17 & 75.38 & 11.24 & 2.13 & 95.82 & \textcolor{red}{3.65} & \textcolor{red}{0.62} \\ 
\hline
Co-train\cite{huang2020multi} & \textcolor{red}{87.40} & 11.74 & 3.01 & \textcolor{blue}{85.04} & 10.34 & 4.07 & 88.53 & 52.43 & 8.82 & 75.93 & 10.81 & 1.90 & 95.82 & 5.18 & 0.77\\ 
\hline
DoDNet\cite{zhang2021dodnet} & 86.63 & 14.49 & 2.66 & 84.89 & 18.77 & 4.98 & 89.94 & 22.01 & 4.50 & \textcolor{blue}{76.51} & \textcolor{blue}{7.34} & \textcolor{blue}{1.61} & 94.45 & 20.44 & 3.58\\ 
\hline
COSST (ours) & 86.05 & \textcolor{red}{9.50} & 2.44 & \textcolor{red}{86.94} & \textcolor{red}{7.78} & \textcolor{red}{1.62} & \textcolor{red}{92.34} & \textcolor{red}{4.13} & \textcolor{red}{1.25} & \textcolor{red}{77.17} & \textcolor{red}{5.90} & \textcolor{red}{1.36} & \textcolor{red}{96.28} & \textcolor{blue}{3.86} & \textcolor{blue}{0.71} \\ 
\hline
\end{tabular}

\begin{tabular}{|D|C|C|C|C|C|C|C|C|C|C|C|C|C|C|C|} 
\hline\hline
\multirow{2}{*}{Methods} & \multicolumn{3}{c|}{Left kidney (F2)} & \multicolumn{3}{c|}{Right kidney (F2)} & \multicolumn{3}{c|}{Spleen (F2)} & \multicolumn{3}{c|}{Pancreas (F2)} & \multicolumn{3}{c|}{Liver (F2)} \\ 
\cline{2-16}
 & DSC &HD95 & ASD & DSC &HD95 & ASD & DSC &HD95 & ASD & DSC &HD95 & ASD & DSC &HD95 & ASD \\ 
\hline
Multi-Nets & 87.10 & 17.26 & 3.66 & 83.59 & 13.57 & 2.41 & 81.62 & 38.49 & 7.15 & 66.72 & 29.23 & 5.48 & 92.23 & 7.96 & 1.40\\ 
\hline
TAL\cite{fang2020multi} & 88.13 & 8.98 & 1.40 & 85.52 & 9.33 & \textcolor{blue}{1.56} & 87.36 & 13.24 & 2.73 & 67.87 & 16.99 & 3.24 & 93.47 & 6.50 & \textcolor{blue}{1.33} \\ 
\hline
ME\cite{shi2021marginal} & 86.75 & 9.90 & 3.98 & 88.03 & 13.91 & 3.84 & \textcolor{blue}{89.35} & \textcolor{blue}{10.01} & \textcolor{blue}{2.73} & 70.33 & 25.30 & 6.74 & 93.40 & 10.80 & 2.92\\ 
\hline
PLT\cite{liu2022universal} & \textcolor{red}{89.63} & \textcolor{blue}{8.72} & \textcolor{blue}{1.27} & \textcolor{blue}{88.18} & \textcolor{blue}{8.70} & 1.88 & 88.65 & 38.60 & 5.10 & \textcolor{blue}{72.40} & \textcolor{blue}{11.00} & \textcolor{blue}{2.03} & \textcolor{blue}{93.51} & \textcolor{blue}{6.26} & \textcolor{red}{1.12}\\ 
\hline
Co-train\cite{huang2020multi} & 88.71 & 11.99 & 1.67 & 87.75 & 11.17 & 2.09 & 88.56 & 67.60 & 11.33 & 71.61 & 17.29 & 2.71 & 93.22 & 6.49 & 1.10\\ 
\hline
DoDNet\cite{zhang2021dodnet} & 88.26 & 16.81 & 2.27 & 87.71 & 22.55 & 3.65 & 88.13 & 28.23 & 5.32 & 67.73 & 12.72 & 2.32 & 92.96 & 10.50 & 2.28 \\ 
\hline
COSST (ours) & \textcolor{blue}{89.40} & \textcolor{red}{7.75} & \textcolor{red}{0.98} & \textcolor{red}{88.96} & \textcolor{red}{6.75} & \textcolor{red}{0.95} & \textcolor{red}{90.64} & \textcolor{red}{4.14} & \textcolor{red}{0.96} & \textcolor{red}{73.46} & \textcolor{red}{7.49} & \textcolor{red}{1.53} & \textcolor{red}{94.40} & \textcolor{red}{5.32} & 1.69\\ 
\hline
\end{tabular}

\begin{tabular}{|D|>{\columncolor{gray!10}}C|>{\columncolor{gray!10}}C|>{\columncolor{gray!10}}C|>{\columncolor{gray!10}}C|>{\columncolor{gray!10}}C|>{\columncolor{gray!10}}C|>{\columncolor{gray!10}}C|>{\columncolor{gray!10}}C|>{\columncolor{gray!10}}C|>{\columncolor{gray!25}}C|>{\columncolor{gray!25}}C|>{\columncolor{gray!25}}C|CCC} 
\hline\hline
\multirow{2}{*}{Methods} & \multicolumn{3}{>{\columncolor{gray!10}}c|}{Average (P)} & \multicolumn{3}{>{\columncolor{gray!10}}c|}{Average (F1)} & \multicolumn{3}{>{\columncolor{gray!10}}c|}{Average (F2)} & \multicolumn{3}{>{\columncolor{gray!25}}c|}{Average (all)} & \multicolumn{3}{c}{}\\ 
\hhline{~|-|-|-|-|-|-|-|-|-|-|-|-|}
 & DSC &HD95 & ASD & DSC &HD95 & ASD & DSC &HD95 & ASD & DSC &HD95 & ASD & & &\\ 
\hhline{-|-|-|-|-|-|-|-|-|-|-|-|-|}
Multi-Nets & 91.18 & 9.98 & 1.65 & 85.31$^{*}$ & 19.26$^{*}$ & 4.46$^{*}$ & 82.25$^{*}$ & 21.30$^{*}$ & 4.02$^{*}$ & 86.25 & 16.85 & 3.38\\ 
\hhline{-|-|-|-|-|-|-|-|-|-|-|-|-|}
TAL\cite{fang2020multi} & 91.46 & \textcolor{blue}{4.74} & \textcolor{blue}{1.01} & 86.08$^{*}$ & \textcolor{blue}{11.24}$^{*}$ & 2.60$^{*}$ & 84.47$^{*}$ & \textcolor{blue}{11.01}$^{*}$ & \textcolor{blue}{2.05}$^{*}$ & 87.34 & \textcolor{blue}{8.99} & \textcolor{blue}{1.89}\\ 
\hhline{-|-|-|-|-|-|-|-|-|-|-|-|-|}
ME\cite{shi2021marginal} & 91.36 & 5.01 & 1.03 & 86.31$^{*}$ & 13.18$^{*}$ & \textcolor{blue}{2.49}$^{*}$ & 85.57$^{*}$ & 13.98$^{*}$ & 4.04$^{*}$ & 87.75 & 10.72 & 2.52\\ 
\hhline{-|-|-|-|-|-|-|-|-|-|-|-|-|}
PLT\cite{liu2022universal} & \textcolor{blue}{91.70} & 10.71 & 1.56 & 86.12$^{*}$ & 19.35$^{*}$ & 3.71$^{*}$ & \textcolor{blue}{86.47}$^{*}$ & 14.66$^{*}$ & 2.28$^{*}$ & \textcolor{blue}{88.10} & 14.91 & 2.51\\ 
\hhline{-|-|-|-|-|-|-|-|-|-|-|-|-|}
Co-train\cite{huang2020multi} & 91.61 & 12.63 & 1.80 & \textcolor{blue}{86.54}$^{\dagger}$ & 18.10$^{*}$ & 3.71$^{*}$ & 85.7$^{*}$ & 22.91$^{*}$ & 3.78$^{*}$ & 88.04 & 17.88 & 3.10\\ 
\hhline{-|-|-|-|-|-|-|-|-|-|-|-|-|}
DoDNet\cite{zhang2021dodnet} & 91.39 & 11.58 & 2.01 & 86.48$^{*}$ & 16.61$^{*}$ & 3.47$^{*}$ & 83.96$^{*}$ & 18.16$^{*}$ & 3.17$^{*}$ & 87.61 & 15.45 & 2.88 \\ 
\hhline{-|-|-|-|-|-|-|-|-|-|-|-|-|}
COSST (ours) & \textcolor{red}{92.13} & \textcolor{red}{3.57} & \textcolor{red}{0.79} & \textcolor{red}{87.76} & \textcolor{red}{6.23} & \textcolor{red}{1.48} & \textcolor{red}{87.37} & \textcolor{red}{6.29} & \textcolor{red}{1.22} & \textcolor{red}{89.08} & \textcolor{red}{5.36} & \textcolor{red}{1.16}\\ 
\hhline{-|-|-|-|-|-|-|-|-|-|-|-|-|}
\end{tabular}

\end{table*}

\newcolumntype{C}{ >{\centering\arraybackslash} m{0.6cm} }
\newcolumntype{D}{ >{\centering\arraybackslash} m{1.67cm} }

\begin{table*}
\centering
\caption{Performance comparison (DSC, \%, higher is better; ASD, mm, lower is better) of partially labeled segmentation methods for \textbf{task2} on our private \underline{bowel} datasets. Red and blue denote the \textcolor{red}{best} and \textcolor{blue}{second best} scores. $\dagger$ and $*$ represent p-value <0.05 and <0.01 using Wilcoxon signed-rank test; otherwise, p-value >0.05. Upper bound: model trained with fully-annotated datasets.}
\label{tab5}
\begin{tabular}{|D|C|C|C|C|C|C|C|C|C|C|>{\columncolor{gray!25}}C|>{\columncolor{gray!25}}C|} 
\hline
\multirow{2}{*}{Methods} & \multicolumn{2}{c|}{Duodenum} & \multicolumn{2}{c|}{Small Bowel} & \multicolumn{2}{c|}{Large Bowel} & \multicolumn{2}{c|}{Sigmoid} & \multicolumn{2}{c|}{Rectum} & \multicolumn{2}{>{\columncolor{gray!25}}c|}{Average} \\ 
\hhline{~|-|-|-|-|-|-|-|-|-|-|-|-|}
 & DSC & ASD & DSC & ASD & DSC & ASD & DSC & ASD & DSC & ASD & DSC & ASD \\ 
\hhline{-|-|-|-|-|-|-|-|-|-|-|-|-|}
Multi-Nets & 70.57 & 2.82 & 83.70$^{*}$ & 2.70$^{*}$ & 81.83$^{\dagger}$ & 3.80$^{\dagger}$ & 66.12$^{*}$ & 7.66 & 73.31 & 4.58 & 75.11 & 4.32 \\ 
\hhline{-|-|-|-|-|-|-|-|-|-|-|-|-|}
TAL\cite{fang2020multi} & 63.82$^{*}$ & 3.20$^{*}$ & 85.12$^{*}$ & 2.38$^{\dagger}$ & 84.05$^{*}$ & 3.32$^{*}$ & 68.13$^{*}$ & 6.52 & 74.46 & \textcolor{blue}{3.42} & 75.12 & 3.76 \\ 
\hhline{-|-|-|-|-|-|-|-|-|-|-|-|-|}
ME\cite{shi2021marginal} & 70.59$^{*}$ & 2.88$^{*}$ & \textcolor{blue}{85.80}$^{*}$ & 2.66$^{*}$ & \textcolor{blue}{84.93}$^{*}$ & 3.08$^{*}$ & \textcolor{blue}{70.24}$^{*}$ & 7.46 & 75.60 & 3.88$^{\dagger}$ & 77.43 & 4.00 \\ 
\hhline{-|-|-|-|-|-|-|-|-|-|-|-|-|}
PLT\cite{liu2022universal} & 71.51 & \textcolor{blue}{2.56} & 84.79$^{*}$ & 2.32 & 83.93$^{*}$ & 3.12 & 68.27 & 6.44$^{\dagger}$ & 75.60 & \textcolor{red}{3.28}$^{*}$ & 76.82 & 3.54 \\ 
\hhline{-|-|-|-|-|-|-|-|-|-|-|-|-|}
Co-train\cite{huang2020multi} & \textcolor{red}{71.95} & \textcolor{red}{2.52}$^{\dagger}$ & 85.54$^{*}$ & \textcolor{red}{2.16} & 84.72$^{\dagger}$ & \textcolor{blue}{3.04} & 69.94 & \textcolor{blue}{5.94} & \textcolor{blue}{76.06} & 3.88 & \textcolor{blue}{77.64} & \textcolor{blue}{3.52} \\ 
\hhline{-|-|-|-|-|-|-|-|-|-|-|-|-|}
MS-KD\cite{feng2021ms} & 65.16$^{*}$ & 5.02$^{*}$ & 83.06$^{*}$ & 3.54$^{*}$ & 80.76$^{*}$ & 7.08$^{*}$ & 66.06$^{*}$ & 12.70$^{\dagger}$ & 70.19$^{*}$ & 5.42$^{*}$ & 73.04 & 6.76 \\ 
\hhline{-|-|-|-|-|-|-|-|-|-|-|-|-|}
DoDNet\cite{zhang2021dodnet} & 42.97$^{*}$ & 98.44$^{*}$ & 84.39$^{*}$ & 3.00$^{*}$ & 83.28$^{*}$ & 5.50$^{*}$ & 67.16$^{*}$ & 9.46$^{*}$ & 56.06$^{*}$ & 102.84$^{*}$ & 66.77 & 43.84 \\ 
\hhline{-|-|-|-|-|-|-|-|-|-|-|-|-|}
COSST (ours) & \textcolor{blue}{71.94} & 2.74 & \textcolor{red}{86.45} & \textcolor{blue}{2.30} & \textcolor{red}{86.07} & \textcolor{red}{3.02} & \textcolor{red}{70.84} & \textcolor{red}{5.72} & \textcolor{red}{76.06} & 3.60 & \textcolor{red}{78.27} & \textcolor{red}{3.48} \\ 
\hline
\hline
Upper bound & 74.41 & 2.84 & 86.92 & 2.28 & 86.09 & 3.04 & 73.67 & 5.28 & 77.04 & 3.28 & 79.63 & 3.34 \\
\hhline{-|-|-|-|-|-|-|-|-|-|-|-|-|}
\end{tabular}
\end{table*}

\newcolumntype{C}{ >{\centering\arraybackslash} m{0.54cm} }
\newcolumntype{D}{ >{\centering\arraybackslash} m{1.67cm} }

\begin{table*}
\centering
\caption{Performance comparison for \textbf{task3} on our private \underline{pelvic} datasets. Formatting is the same as Table \ref{tab5}. L: left. R: right. Sem.Ves: seminal vesicles}
\label{tab6}
\begin{tabular}
{|D|C|C|C|C|C|C|C|C|C|C|C|C|C|C|>{\columncolor{gray!25}}C|>{\columncolor{gray!25}}C|} 
\hline
\multirow{2}{*}{Methods} & \multicolumn{2}{c|}{Bladder} & \multicolumn{2}{c|}{Prostate} & \multicolumn{2}{c|}{Rectum} & \multicolumn{2}{c|}{Femur (L)} & \multicolumn{2}{c|}{Femur (R)} & \multicolumn{2}{c|}{Sem.Ves} & \multicolumn{2}{c|}{Uterus} & \multicolumn{2}{>{\columncolor{gray!25}}c|}{Average} \\
\hhline{~|-|-|-|-|-|-|-|-|-|-|-|-|-|-|-|-|}
 & DSC & ASD & DSC & ASD & DSC & ASD & DSC & ASD & DSC & ASD & DSC & ASD & DSC & ASD & DSC & ASD \\ 
\hline
Multi-Nets & 87.80$^{*}$ & 6.88$^{*}$ & 77.51 & \textcolor{blue}{2.16} & 83.42$^{\dagger}$ & 2.16 & 93.15$^{*}$ & \textcolor{red}{0.78}$^{*}$ & 93.15$^{*}$ & \textcolor{blue}{0.96} & 71.40$^{\dagger}$ & 1.58$^{\dagger}$ & 78.59 & 3.98 & 83.57 & 2.64\\ 
\hline
TAL\cite{fang2020multi} & 87.82$^{*}$ & 3.60$^{*}$ & 73.88$^{*}$ & 2.30$^{*}$ & 82.06$^{*}$ & 2.28$^{*}$ & 94.78 & 1.26  & 94.53 & 0.98 & 65.25 & 1.90 & 75.37$^{*}$ & \textcolor{blue}{3.74} & 81.96 & 2.30 \\ 
\hline
ME\cite{shi2021marginal} & 88.97$^{*}$ & \textcolor{blue}{1.82}$^{*}$ & \textcolor{blue}{79.56} & 2.28$^{*}$ & 84.62$^{*}$ & 2.76$^{*}$ & 94.11$^{*}$ & 1.08$^{*}$  & 93.28$^{*}$ & 1.24$^{*}$ & \textcolor{blue}{73.64} & 1.50$^{*}$ & \textcolor{red}{79.65} & 3.96 & \textcolor{blue}{84.83} & \textcolor{blue}{2.10}\\ 
\hline
PLT\cite{liu2022universal} & 88.10 & 3.80 & 77.31 & 6.06 & 83.34 & 2.46 & 93.80 & 2.92 & 93.45 & 2.62$^{*}$ & 73.34 & \textcolor{blue}{1.40} & 75.79 & 4.52 & 83.59 & 3.40\\ 
\hline
Co-train\cite{huang2020multi} & \textcolor{red}{89.71} & 3.10 & 79.20 & \textcolor{red}{2.10} & \textcolor{blue}{85.84} & \textcolor{blue}{2.12} & \textcolor{blue}{95.01} & 0.86 & \textcolor{red}{94.62} & \textcolor{red}{0.90}$^{*}$ & 72.58 & 1.42 & 73.91 & 4.44 & 84.41 & 2.14\\ 
\hline
MS-KD\cite{feng2021ms} & 78.55$^{*}$ & 7.86 & 69.84 & 3.14 & 77.98 & 9.36 & 91.90$^{*}$ & 4.82$^{*}$ & 92.65$^{*}$ & 2.70$^{*}$ & 46.02 & 2.64 & 74.16 & 7.34 & 75.87 & 5.40\\ 
\hline
DoDNet\cite{zhang2021dodnet} & 88.09$^{*}$ & 5.90$^{*}$ & 77.55$^{*}$ & 4.26$^{*}$ & 85.24$^{*}$ & 2.34$^{*}$ & 62.44$^{*}$ & 90.32$^{*}$ & 63.24$^{*}$ & 88.64$^{*}$ & \textcolor{red}{73.74}$^{*}$ & 3.90$^{*}$ & \textcolor{blue}{79.35} & 4.16$^{*}$ & 75.67 & 28.50\\ 
\hline
COSST (ours) & \textcolor{blue}{89.43} & \textcolor{red}{1.54} & \textcolor{red}{79.64} & 2.20 & \textcolor{red}{85.84} & \textcolor{red}{2.06} & \textcolor{red}{95.14} & \textcolor{blue}{0.82}  & \textcolor{blue}{94.59} & 0.96 & 73.50 & \textcolor{red}{1.40} & 78.99 & \textcolor{red}{3.72} & \textcolor{red}{85.30} & \textcolor{red}{1.82} \\
\hline
\end{tabular}
\end{table*}

\newcolumntype{C}{ >{\centering\arraybackslash} m{0.54cm} }
\newcolumntype{D}{ >{\centering\arraybackslash} m{1.67cm} }

\begin{table*}
\centering
\caption{Performance comparison for \textbf{task4} on our private \underline{eye} datasets. Formatting is the same as Table \ref{tab5}. L: left. R: right. ON: optic nerve.}
\label{tab7}
\begin{tabular}{|D|C|C|C|C|C|C|C|C|C|C|C|C|C|C|>{\columncolor{gray!25}}C|>{\columncolor{gray!25}}C|} 
\hline
\multirow{2}{*}{Methods} & \multicolumn{2}{c|}{Chiasm} & \multicolumn{2}{c|}{ON (L)} & \multicolumn{2}{c|}{ON (R)} & \multicolumn{2}{c|}{Len (L)} & \multicolumn{2}{c|}{Len (R)} & \multicolumn{2}{c|}{Eyeball (L)} & \multicolumn{2}{c|}{Eyeball (R)} & \multicolumn{2}{>{\columncolor{gray!25}}c|}{Average}\\ 
\hhline{~|-|-|-|-|-|-|-|-|-|-|-|-|-|-|-|-|}
 & DSC & ASD & DSC & ASD & DSC & ASD & DSC & ASD  & DSC & ASD & DSC & ASD & DSC & ASD & DSC & ASD\\ 
\hline
Multi-Nets & 48.76$^{\dagger}$ & \textcolor{blue}{1.24} & 65.54$^{*}$ & 0.82 & 66.55$^{\dagger}$ & 0.82$^{\dagger}$ & 76.66 & 0.56  & 76.53 & 0.49 & 92.57 & \textcolor{red}{0.54} & \textcolor{blue}{92.59} & \textcolor{blue}{0.54} & 74.17 & 0.72 \\ 
\hline
TAL\cite{fang2020multi} & 44.98$^{*}$ & 1.18 & 61.27$^{*}$ & 0.84 & 62.99$^{*}$ & 0.80 & 62.73$^{*}$ & 0.55$^{*}$  & 70.23$^{*}$ & 0.55$^{*}$ & 91.47$^{*}$ & 0.55$^{*}$ & 91.87$^{*}$ & 0.55$^{*}$ & 69.36 & 0.76\\ 
\hline
ME\cite{shi2021marginal} & 49.56$^{*}$ & 1.29$^{*}$ & 66.38$^{*}$ & 0.86$^{*}$ & 67.21$^{*}$ & 0.83$^{*}$ & \textcolor{red}{77.09} & \textcolor{blue}{0.53}  & 77.70 & 0.50$^{*}$ & 91.86$^{*}$ & 0.74$^{*}$ & 91.82$^{*}$ & 0.55$^{*}$ & 74.52 & 0.77\\ 
\hline
PLT\cite{liu2022universal} & 49.58$^{\dagger}$ & \textcolor{red}{1.21}$^{\dagger}$ & 66.35 & 0.81 & \textcolor{red}{67.06} & 0.79$^{\dagger}$ & \textcolor{blue}{77.07} & 0.55  & \textcolor{blue}{77.79} & 0.50 & \textcolor{red}{92.57}$^{*}$ & \textcolor{blue}{0.55}$^{\dagger}$ & 92.59 & 0.55 & 74.72 & 0.71\\ 
\hline
Co-train\cite{huang2020multi} & \textcolor{blue}{49.60} & 1.25 & \textcolor{blue}{66.57} & \textcolor{red}{0.80} & 67.45 & \textcolor{red}{0.78} & 76.84 & 0.55 & 77.41 & \textcolor{red}{0.47} & \textcolor{blue}{92.52}$^{\dagger}$ & 0.55$^{*}$ & \textcolor{red}{92.78} & \textcolor{red}{0.52}$^{*}$ & \textcolor{blue}{74.74} & \textcolor{red}{0.70}\\ 
\hline
MS-KD\cite{feng2021ms} & 47.07$^{*}$ & 1.87$^{*}$ & 57.63$^{*}$ & 1.50$^{*}$ & 58.17$^{*}$ & 1.45$^{*}$ & 76.37 & 0.81$^{\dagger}$ & 77.25 & 0.55$^{\dagger}$ & 92.17 & 0.55 & 91.80$^{*}$ & 0.73$^{*}$ & 71.50 & 1.07\\ 
\hline
DoDNet\cite{zhang2021dodnet} & 47.86 & 1.32 & 41.40$^{*}$ & 17.47$^{*}$ & 43.54$^{*}$ & 17.32$^{*}$ & 43.91$^{*}$ & 33.23$^{*}$  & 51.47$^{*}$ & 28.43$^{*}$ & 60.54$^{*}$ & 26.52$^{*}$ & 61.80$^{*}$ & 26.16$^{*}$ & 50.07 & 21.49\\ 
\hline
COSST (ours) & \textcolor{red}{50.55} & 1.26 & \textcolor{red}{67.17} & \textcolor{blue}{0.81} & \textcolor{blue}{67.89} & \textcolor{blue}{0.79} & 76.89 & \textcolor{red}{0.53} & \textcolor{red}{77.99} & \textcolor{blue}{0.48} & 92.31 & 0.58 & 92.52 & 0.56 & \textcolor{red}{75.05} & \textcolor{blue}{0.71}\\
\hline
\end{tabular}
\end{table*}

\subsection{Comparison With State-of-the-Art Methods}
We compare our proposed COSST against seven state-of-the-art approaches that also address the partial-label segmentation problem. The compared methods are (1) individual networks trained on each partially labeled dataset (Multi-Nets), (2) two methods that utilize only the ground truth-based supervision: target adaptive loss \cite{fang2020multi} and marginal and exclusive loss \cite{shi2021marginal} (denoted as TAL and ME), (3) three methods that employ pseudo label supervision: pseudo label training \cite{liu2022universal}, Co-training of weight-averaged models \cite{huang2020multi}, a multi-teacher single-student knowledge distillation framework that exploits soft pseudo labels (denoted as PLT, Co-training, and MS-KD), (4) DoDNet \cite{zhang2021dodnet}: a state-of-the-art conditioned network. To ensure fair comparison, we use the same backbone architecture and training strategies for all compared methods.

\begin{figure*}[t]
\includegraphics[width=1\textwidth]{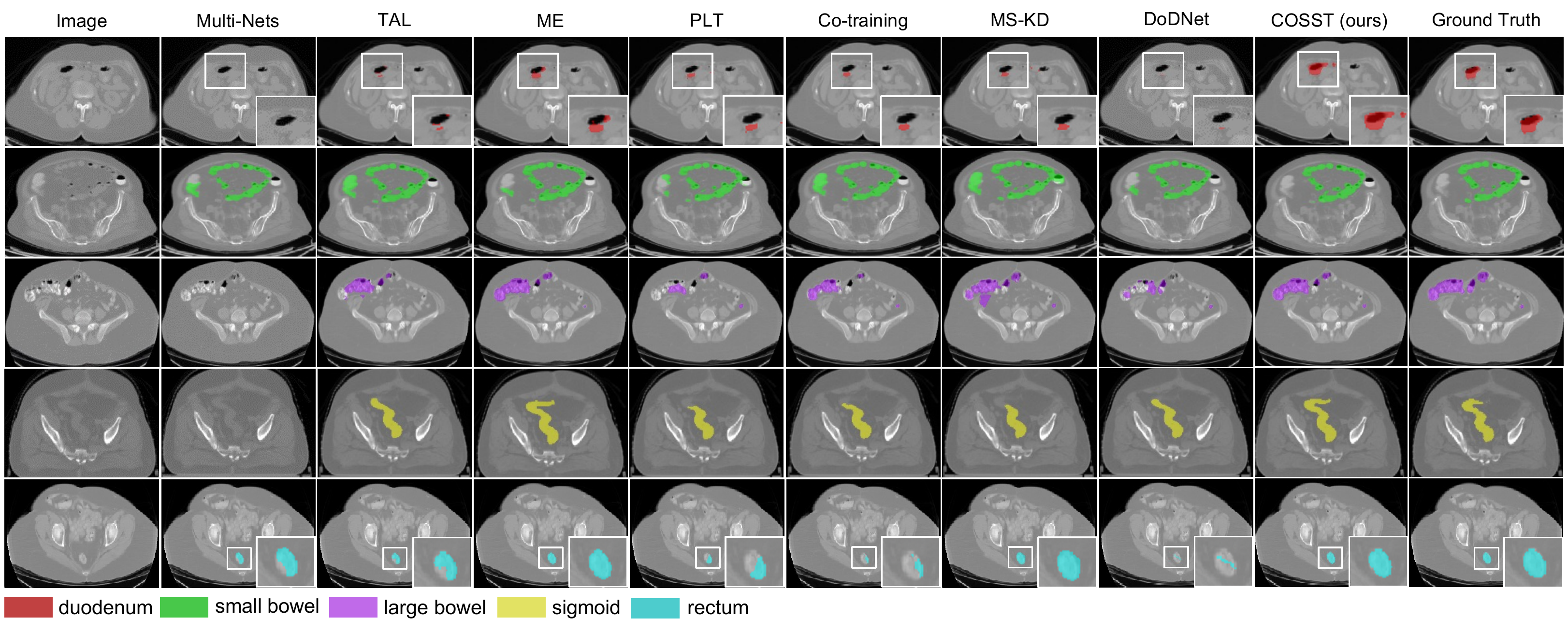}
\centering
\caption{Qualitative comparisons between our proposed COSST and other partial-label segmentation methods on \underline{bowel} datasets for \textbf{task2}.}
\label{fig5}
\end{figure*} 

\begin{figure*}[t]
\includegraphics[width=1\textwidth]{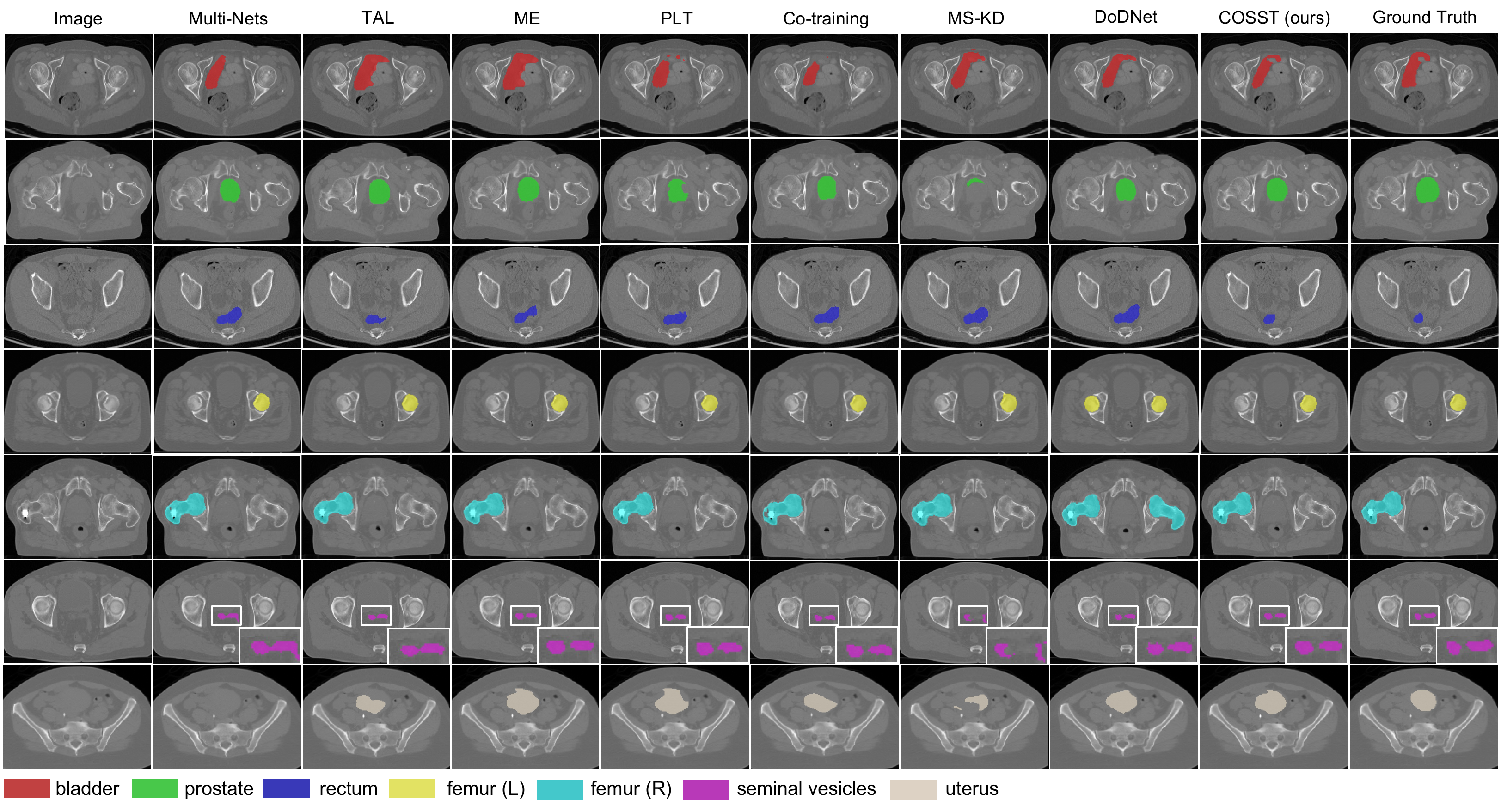}
\centering
\caption{Qualitative comparisons between our proposed COSST and other partial-label segmentation methods on \underline{pelvic} datasets for \textbf{task3}.}
\label{fig6}
\end{figure*} 

\begin{figure*}[t]
\includegraphics[width=1\textwidth]{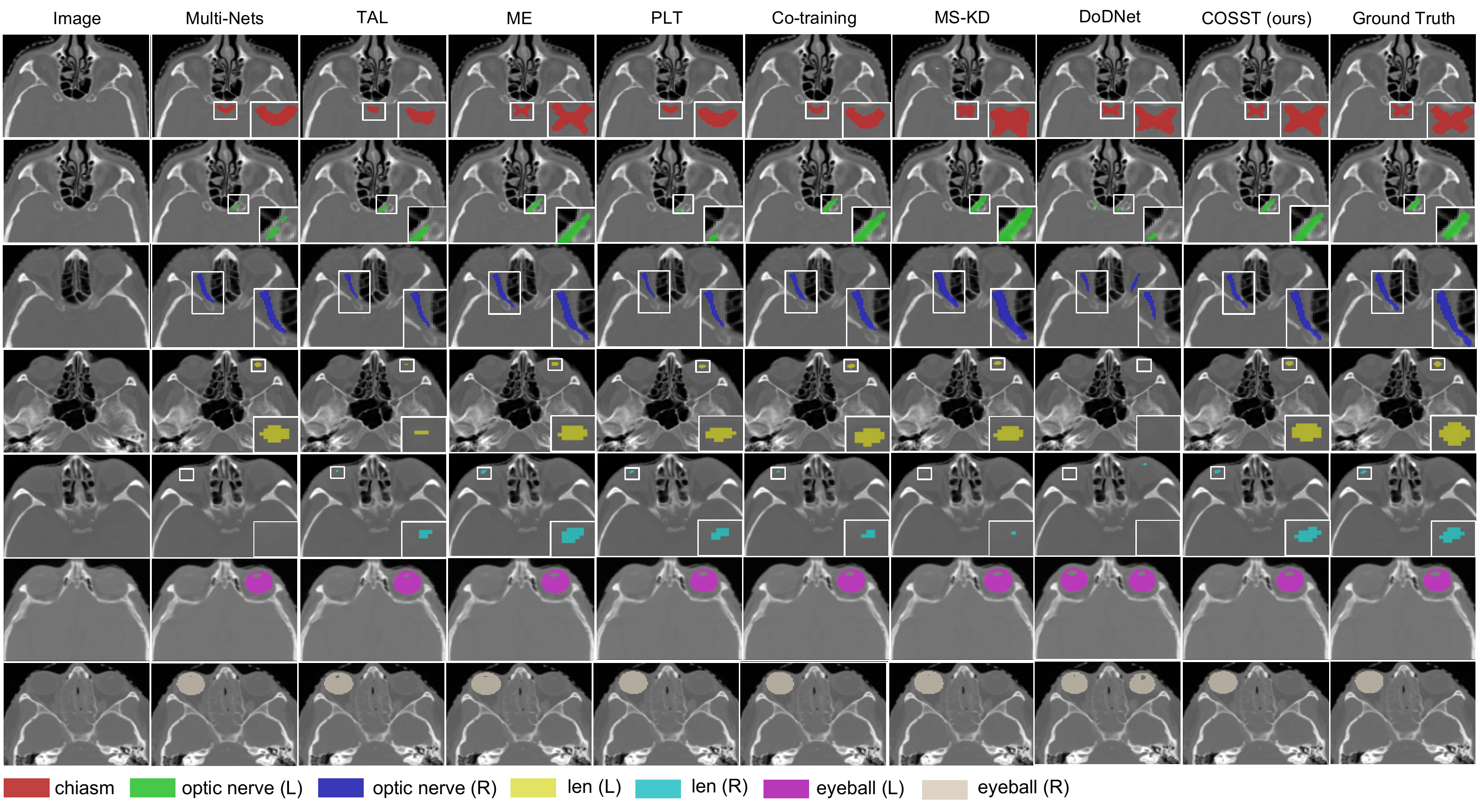}
\centering
\caption{Qualitative comparisons between our proposed COSST and other partial-label segmentation methods on \underline{eye} datasets for \textbf{task4}.}
\label{fig7}
\end{figure*} 

In Tab. \ref{tab4}, we show the segmentation performance of the compared methods for the public datasets in \textbf{task1}. First, we observe that it is more beneficial to learn a single unified model from partially labeled datasets than the baseline Multi-Nets, especially for the organs whose training set is small, e.g., spleen (N=20). For example, by comparing the Multi-Nets to the proposed COSST, we can observe that the average Dice score improves from 86.25\% to 89.08\%, the average HD95 improves from 16.85 to 5.36, and the average ASD improves from 3.38 to 1.16. By comparing the segmentation performance across three evaluated datasets, we notice that the segmentation performance of all compared methods decreases from the held-out testing set to the external datasets, e.g., the average Dice score of COSST drops from 92.13\% to 87.76\% and 87.37\% for BTCV and AMOS2022, respectively. However, we find that COSST consistently yields more reliable segmentation results than other methods under the domain shift. For instance, by the comparing the COSST to the second best approach, we show that the segmentation results of COSST have much better boundary matching, i.e., 6.23 vs. 11.24 and 6.29 vs. 11.01 respectively for BTCV and AMOS2022. For ground truth based supervision methods, we observe that TAL and ME achieve comparable results in Dice scores but TAL slightly outperforms ME in HD95 and ASD. In addition, we notice that the pseudo-label based methods such as PLT and co-training achieve better segmentation performance than the non-pseudo label based methods, i.e., TAL and ME, demonstrating the importance of the pseudo label learning. Furthermore, comparing the pseudo label based methods (e.g., PLT) to our proposed COSST, we observe that COSST not only achieves overall higher Dice scores, but also much better performance in distance-based metrics HD95 and ASD (ASD: 1.16 vs. 2.51, and HD95: 5.36 vs. 14.91), indicating the effectiveness of our strategy for pseudo label learning, i.e., pseudo label filtering with self-training. Overall, COSST achieves consistent superior segmentation performance than all compared methods on all three evaluated CT datasets, especially in the distance-based metrics HD95 and ASD.

Tab. \ref{tab5}, \ref{tab6}, and \ref{tab7} tabulate the segmentation performance on our private datasets for \textbf{task2}, \textbf{task3}, and \textbf{task4}, respectively. Our results reveal that most partial-label segmentation approaches outperform the baseline Multi-Nets, demonstrating the benefits of training a single network on the union of partially labeled datasets. For the ground truth-based supervision methods, we observe that ME consistently outperforms TAL on all three segmentation tasks. For the pseudo label based methods, Co-training achieves consistent better performance than PLT. The performance of MS-KD does not appear competitive as it is even worse than the baseline Multi-Nets. The conditioned network DoDNet achieves sub-optimal results in our experiments, especially for the small organs in bowel datasets, i.e., duodenum and rectum. However, it achieves superior performance on the gender-specific organs such as seminal vesicles and uterus in pelvic datasets (Tab. \ref{tab6}). In addition, we notice it fails to distinguish the left and right labels for symmetric organs, such as femurs and optic nerves. Lastly, the proposed COSST achieves the highest overall segmentation performance among the competing partial-label segmentation methods on all three segmentation tasks (except the second best ASD on eye dataset). Furthermore, in Tab. \ref{tab5}, our results show that the performance achieved by COSST is comparable to the upper bound, i.e., the network trained with fully-annotated datasets. Especially, COSST achieves significant improvements on complex structures such as small bowel and large bowel. Qualitatively, we also observe in Fig. \ref{fig5} that COSST provides more reasonable segmentation than other partial-label segmentation approaches on bowel datasets.

\subsection{Ablation Studies}
We conduct ablation studies on the \textbf{task2} (\underline{bowel datasets}) to investigate several important questions regarding our method.

\begin{figure}[t]
\centerline{\includegraphics[width=1\columnwidth]{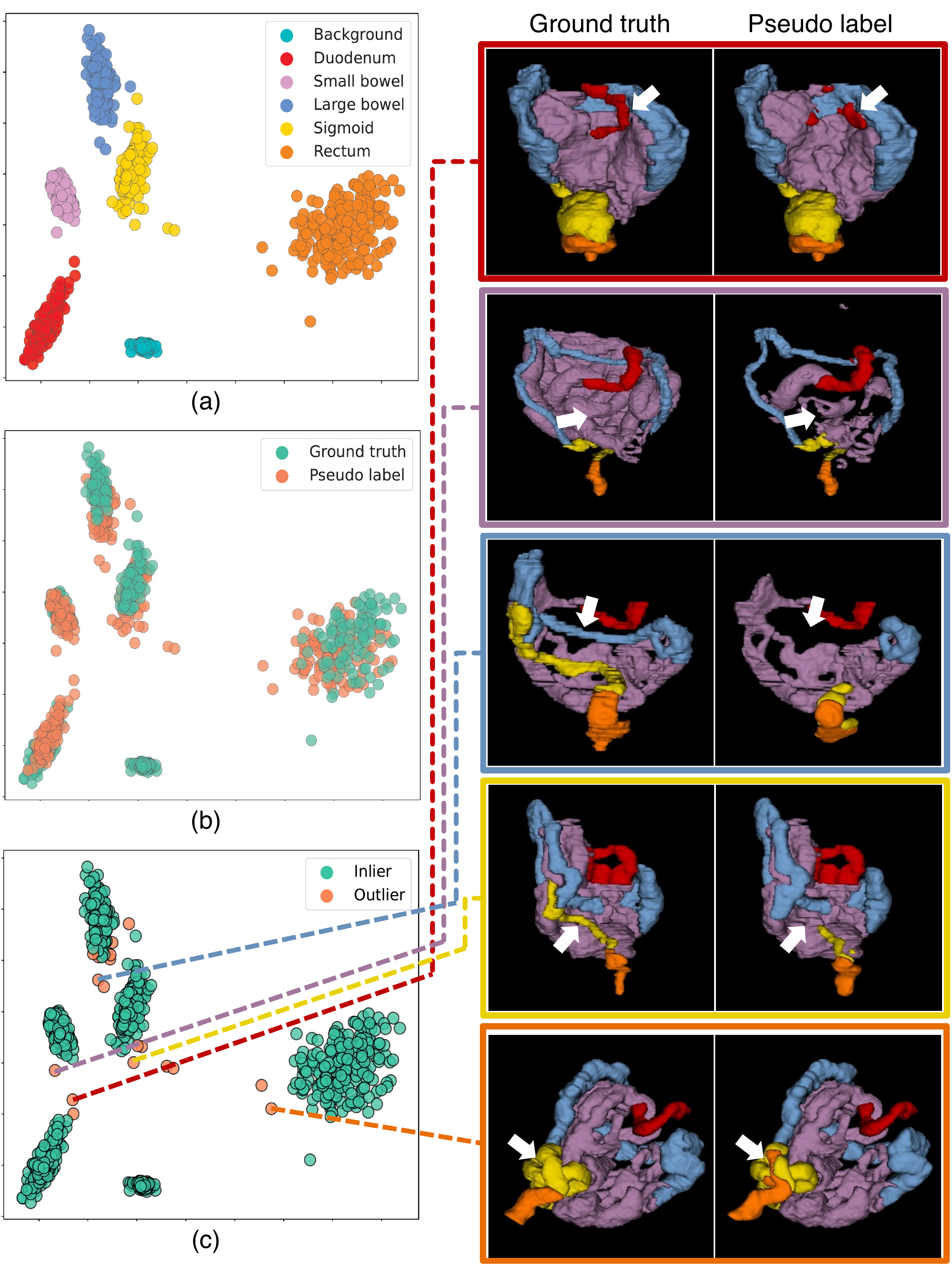}}
    \caption{Left panel: Visualization of the organ-wise feature representations with 2D PCA on the bowel datasets, color coded by (a) organ names, (b) ground truth or pseudo label, and (c) detected inlier or outlier. Right panel: Qualitative comparison between the detected unreliable pseudo labels (outliers) and the ground truth for each organ. Major differences are marked by white arrows.}
    \label{fig8}
\end{figure}

\subsubsection{Effectiveness of Pseudo Label Assessment}
\begin{figure}[t]
\centerline{\includegraphics[width=1\columnwidth]{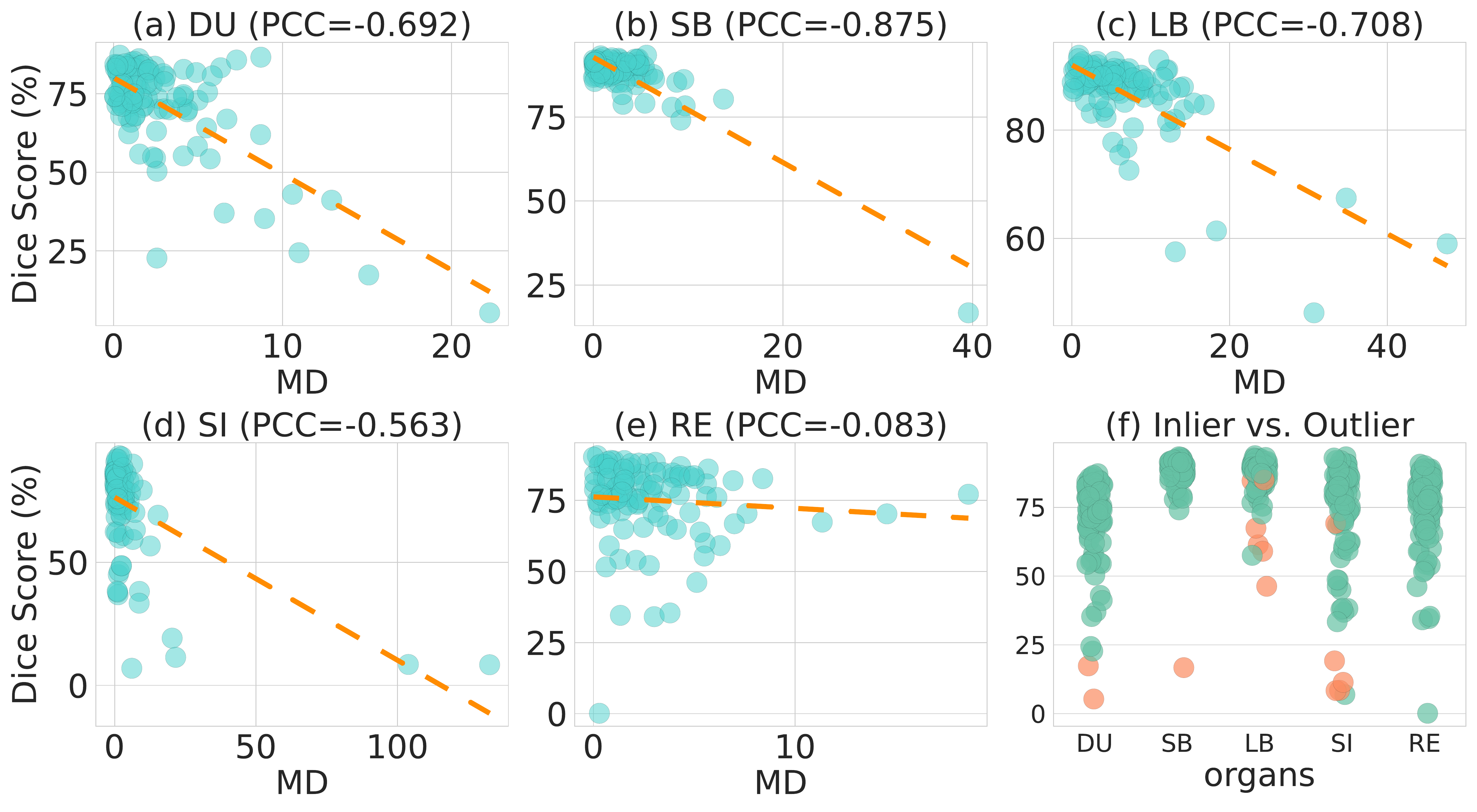}}
\caption{(a)-(e) display the Dice scores of pseudo labels vs. their corresponding assessment metric, i.e., Mahalanobis distance (MD). Strong correlations are observed for DU, SB and LB, moderate correlation for SI, and weak correlation for RE. The dashed line represents linear regression model fit. (f) The detected inliers (green) and outliers (orange) for each organ. Most detected outliers are among the pseudo labels with the lowest Dice scores across the entire distribution. DU: duodenum, SB: small bowel, LB: large bowel, SI: sigmoid, RE: rectum.}
\label{fig9}
\end{figure}

In this section, we evaluate the effectiveness of our pseudo label assessment strategy. As shown in Fig. \ref{fig8}, we first visualize the organ-wise feature representations obtained by Eq. 3 using 2D PCA on the left panel. In Fig. \ref{fig8} (a), we observe that most feature vectors belonging to the same organ are well clustered. In Fig. \ref{fig8} (b), for each organ, the ground truth distribution is highly entangled with the pseudo label distribution. In Fig. \ref{fig8} (c), we visualize the detected outliers (unreliable pseudo labels) identified by our pseudo label assessment strategy. Given the additional annotations of the initially unlabeled organs, we comprehensively evaluate the quality of the pseudo labels that are identified as outliers. On the right panel of Fig. \ref{fig8}, our qualitative comparison shows that the detected pseudo labels have significant differences in shape compared to the ground truth. For quantitative comparison, we compute the Dice scores of all pseudo labels against ground truth and calculate the Pearson Correlation Coefficient (PCC) between the our assessment metric, i.e., Mahalanobis distance, and the Dice scores. As shown in Fig. \ref{fig9} (a)-(e), we observe strong correlations for most organs, i.e., duodenum, small bowel and large bowel, moderate correlation for sigmoid, but weak correlation for rectum. The underlying reason for the weak correlation of rectum may be that the shape of rectum is relatively small and thus more sensitive to shape variations. As shown in Fig. \ref{fig8} (a), compared to other organs, rectum (denoted by orange dots) has a less compact cluster in latent space, which makes it more difficult to yield a high correlation between the quality of pseudo labels and the distance in latent space. Lastly, in Fig. \ref{fig9} (f), we can clearly see that most detected outliers are among the pseudo labels with the lowest Dice scores across the entire distribution, further verifying the effectiveness of our strategy. 

\subsubsection{Effectiveness of Pseudo Label Filtering}
In this section, we investigate the effectiveness of the different pseudo label filtering schemes for self-training. Specifically, we compare four schemes including (1) no filtering: pseudo labels are used without quality control, (2) image-level filtering (ours), (3) voxel-level filtering which has been shown to effectively denoise the pseudo label masks on voxel-level\cite{zhang2021prototypical}, and (4) the combination of image-level and voxel-level filtering. We report the average Dice scores and ASD of all organs for comparison, as shown in Tab. \ref{tab8}. Our observations are as follows. First, even with no filtering, self-training with the plain pseudo labels has already improved the performance of the initial unified model (row 1 vs. row 2), demonstrating that both the pseudo label supervision can be used for free performance boost and is complementary to the ground truth-based supervision. Second, self-training performance can be further improved by image-level pseudo label filtering, especially the ASD (row 2 vs. row 3), suggesting that the unreliable pseudo labels may have limited the model performance. Lastly, our experiments show that the voxel-level filtering scheme does not enhance the self-training performance for our specific task (row 4 and 5). This indicates that the noisy pseudo labels may not be reliably fixed via voxel-level denoising and they should rather be entirely excluded from training.

\begin{table}[t]
\centering
\caption{Performance on bowel datasets with different pseudo label filtering schemes. COSST (row 3) exploits pseudo labels for training and \underline{only} the image-level pseudo label filtering is applied. The average DSC and ASSD are reported.}
\label{tab8}
\begin{tabular}{|c|c|c|c|c|} 
\hline
pseudo label & image-level & voxel-level & DSC (\%) & ASSD (mm) \\ 
\hline
 & & & 77.43 & 4.00 \\ 
\hline
\cmark & & & 77.85 & 3.96 \\ 
\hline
\cmark & \cmark &  & \textbf{78.27} & \textbf{3.48} \\ 
\hline
 \cmark & & \cmark & 77.74 & 3.98 \\ 
\hline
\cmark & \cmark & \cmark & 77.91 & 3.80 \\
\hline
\end{tabular}
\end{table}

\subsubsection{Impact of Training Data Size}
\begin{figure}[h]
    \centerline{\includegraphics[width=1\columnwidth]{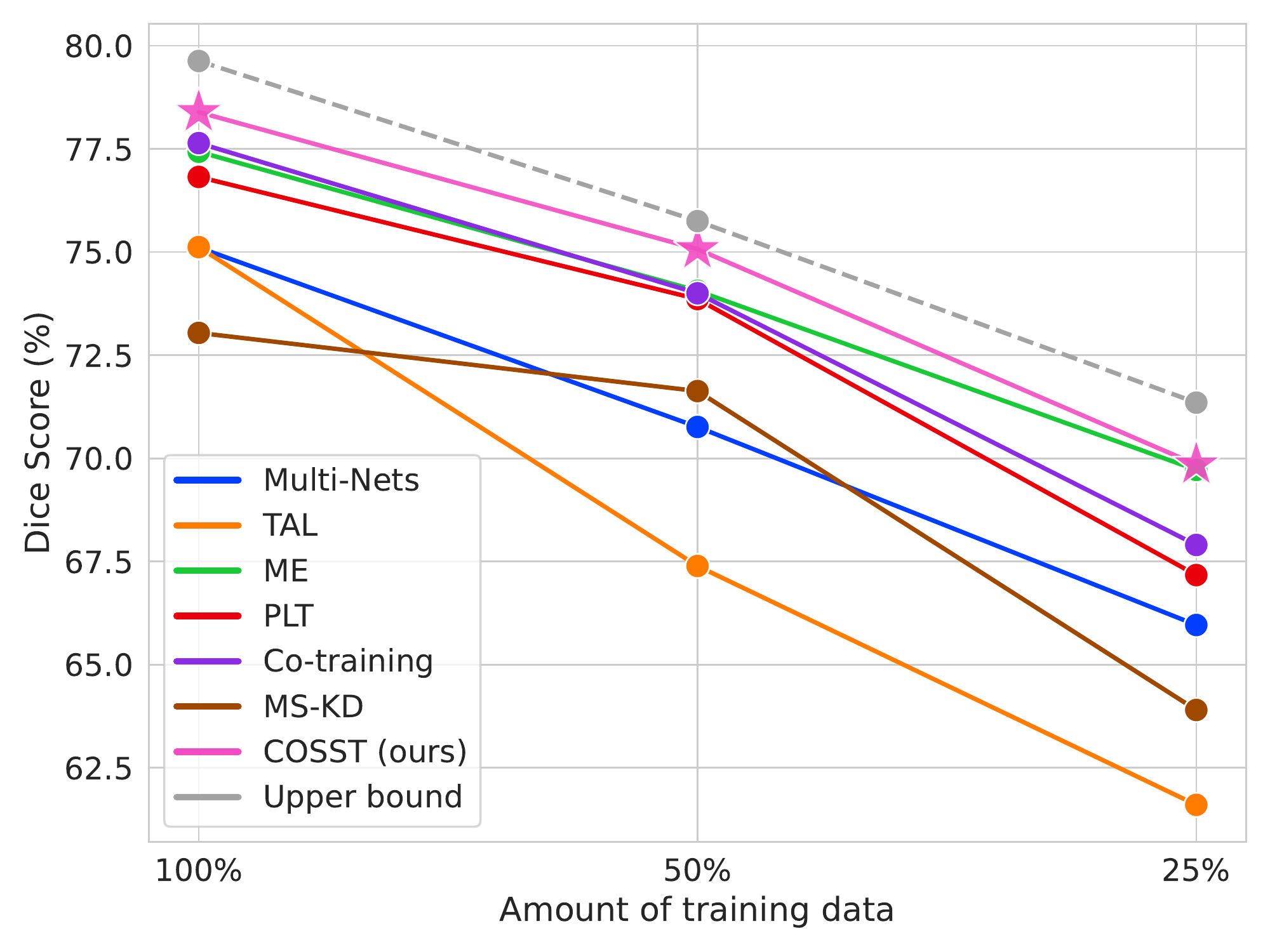}}
    \caption{Performance of partial-label segmentation methods with different training data sizes. Our proposed COSST consistently achieves superior performance when the training data size varies.}
    \label{fig10}
\end{figure}

In this section, we explore the impact of training data size on different partial-label segmentation methods. Specifically, we additionally train all competing methods using only 50\% and 25\% of training data, simulating the scenarios where the size of partially labeled datasets is more limited. As shown in Fig. \ref{fig10}, we observe that the top three benchmark methods, i.e., ME, PLT, and Co-training, achieve comparable performance with 100\% and 50\% of training data, while ME outperforms the other two by a large margin at 25\%. This suggests that the pseudo label based approaches, such as PLT and Co-training, may yield sub-optimal performance in low-data scenarios if the noisy pseudo labels are not removed. It can also be observed that at 25\% training data, COSST achieves slightly better performance than ME. The underlying reason is that, the model trained with only 25\% training data cannot achieve very satisfactory performance and thus the pseudo labels at low-data regime are less reliable, limiting the benefit from pseudo label training. Due to the pseudo label filtering mechanism, our approach does not suffer from performance degradation as PLT and Co-training and can be slightly better than ME at 25\%. In summary, compared to the top-performing benchmark methods, our COSST is more robust to different training data sizes and stands out as a better option given a new partial-label segmentation task.

\subsubsection{Impact of the Threshold for Outlier Detection}
We conduct experiments to explore the impact of different threshold values for outlier detection using Mahalanobis distance, including 0.999, 0.99 and 0.95. First, as shown in Fig. \ref{fig10}, we visualize the inlier vs. outlier plots over three threshold values for Mahalanobis distance. We can observe that more data points are considered as outliers as the threshold value decreases. However, we notice that a low threshold such as 0.95, though removing many poor pseudo labels, may also remove some pseudo labels with reasonable dice scores. This finding aligns with the results in Fig. \ref{fig9}, i.e., the correlation between the Mahalanobis distance and the actual Dice scores is not perfect. To further investigate the impact of thresholds on the segmentation performance, we train the second stage of COSST with different sets of pseudo labels filtered by different thresholds. Our results show that the Dice score on the validation set is improved from 75.45\% (first-stage) to 76.32\%, 76.18\% and 75.81\% for thresholds of 0.999, 0.99, and 0.95, respectively. This result suggests that a conservative threshold is more suitable for our pseudo label filtering method such that the most unreliable pseudo labels can be removed without excluding too many reasonable pseudo labels. Therefore, we empirically set the threshold as 0.999 for all our experiments. Though a fixed threshold may not be optimal for every dataset/task, our results show that a threshold of 0.999, determined based on \textbf{task2}, can be reliably used for other tasks in our experiments.

\begin{figure}[t]
\centerline{\includegraphics[width=1\columnwidth]{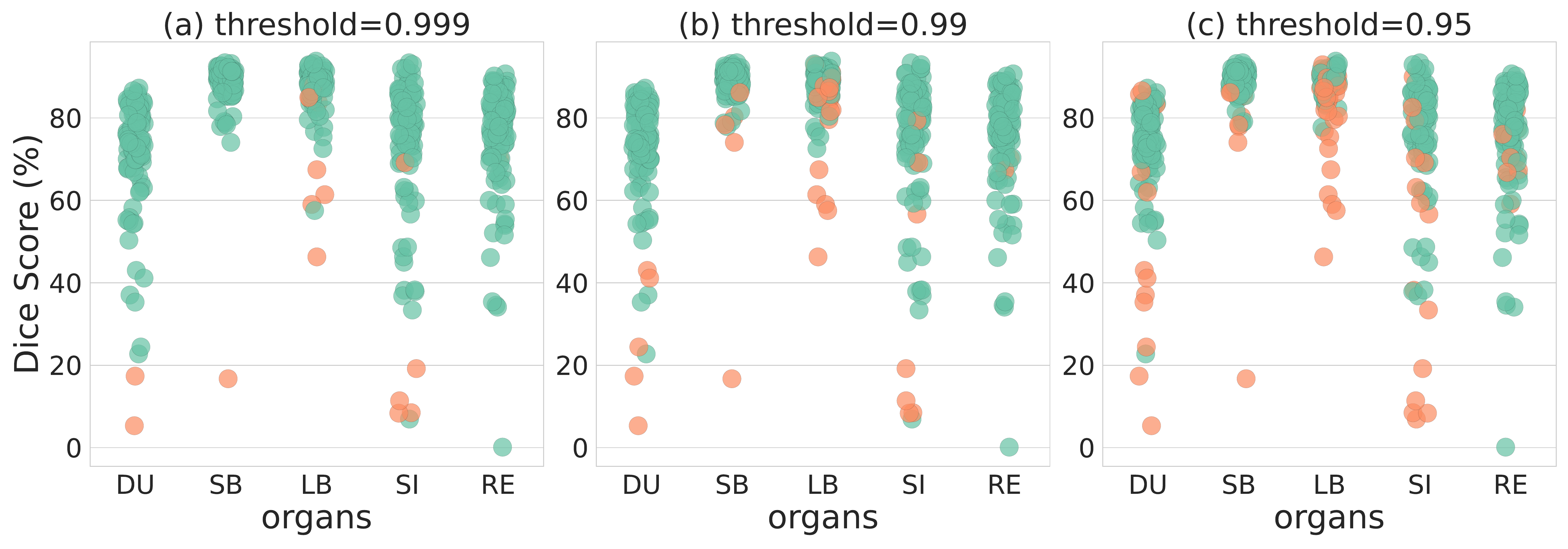}}
    \caption{Outliers (orange) detected by different thresholds for Mahalanobis distance.}
    \label{fig11}
\end{figure}

\begin{figure}[ht]
    \centerline{\includegraphics[width=1\columnwidth]{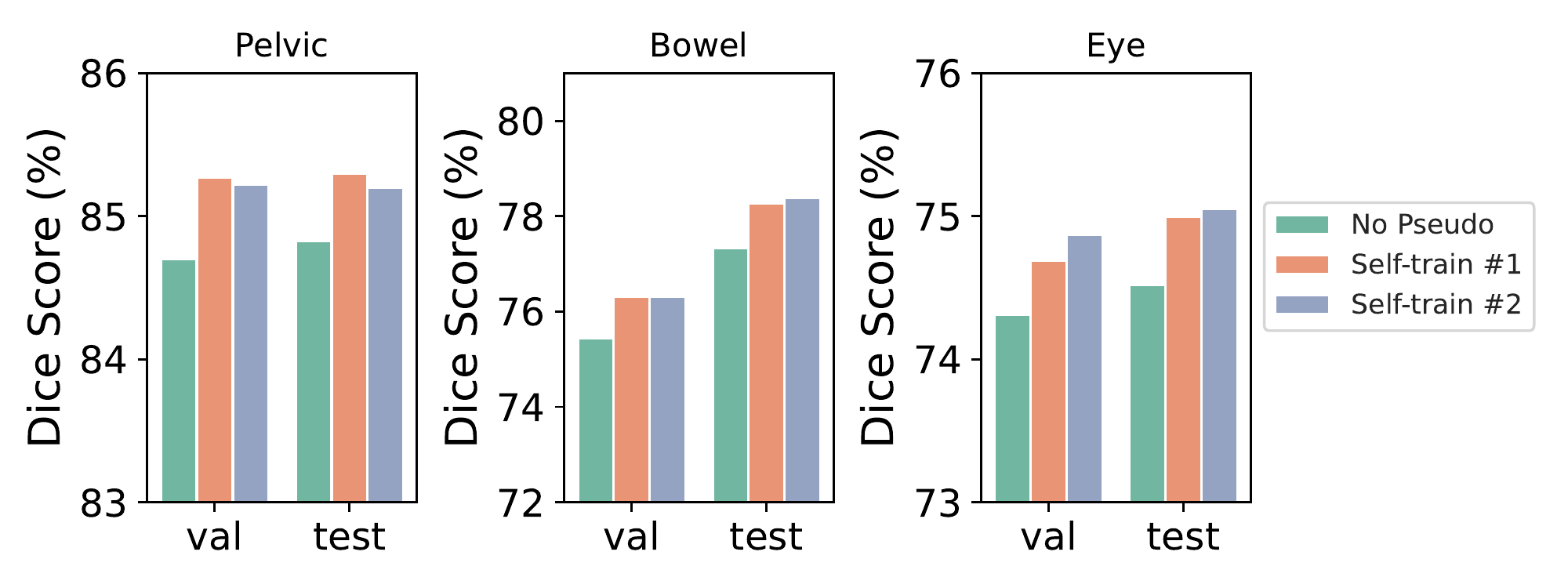}}
    \caption{Performance achieved by different self-training iterations on three segmentation tasks. Self-training mostly converges within one or two iterations and the most significant improvement is observed at the first iteration.}
    \label{fig12}
\end{figure}

\subsubsection{Impact of Self-training Iterations}
We investigate the impact of self-training iterations on model performance on \textbf{task2-4}. As shown in Fig. \ref{fig12}, we observe that self-training typically converges within one or two iterations and the most significant improvement is observed at the first iteration (No Pseudo vs. Self-train \#1). Moreover, in our experiments, we find it effective to use the validation performance to determine when to terminate self-training. However, this finding needs to be interpreted carefully because the data distribution of our validation set may be similar to testing set. Other termination criteria may be used to obtain better self-training results.

\section{Discussion and Conclusion}
\label{sec:VI}
In this study, we systematically investigate the partial-label segmentation problem with both theoretical analyses and empirical evaluations on the prior techniques. We identify three types of supervision signals for partial-label segmentation and show that integration of three supervision signals using self-training and pseudo label filtering can lead to improved performance. In the following sections, we offer a detailed discussion of our observations.

\subsubsection{Unified model vs. Individual models} Our experimental results show that  the unified models that are trained on all partially labeled datasets achieve better segmentation performance than the Multi-Nets that are separately trained on each individual partial-label dataset. The unified models outperform Multi-Nets especially in the distance-based evaluation metrics, indicating that training on more data (even partially labeled) can help improve the reliability of the segmentation results. The consistent outperformance the partial-label learning is observed on all four segmentation tasks in our experiments and our finding also aligns with the results provided by other studies \cite{fang2020multi,shi2021marginal,liu2022universal,huang2020multi,feng2021ms,zhang2021dodnet}. Moreover, our results also show that the superiority of unified models is invariant to the amount of training data used (Fig. \ref{fig10}). Besides the improved performance, unified models are more efficient than Multi-Nets as they can segment all organs of interest simultaneously. By contrast, Multi-Nets needs to combine the results from individual models and thus takes longer inference time. Moreover, Multi-Nets may require extra post-processing steps to address conflicting predictions.

\subsubsection{Analyses of the Prior Techniques} 
In this section, we empirically analyze the benchmark partial-label segmentation methods based on our experimental results. 

First, we compare the two methods that utilize only the ground truth-based supervision signals, namely TAL\cite{fang2020multi} and ME\cite{shi2021marginal}. Compared to TAL which only considers Sup. I, ME imposes an additional supervision (Sup. II) to regularize the predictions of unlabeled organs based on the mutual exclusiveness among organs. In \textbf{task1}, we observe that TAL and ME achieve highly comparable segmentation results in Dice scores but TAL achieves slightly better results in distance-based metrics. However, in \textbf{task2-4}, we find that ME achieves consistent better segmentation results than TAL and can even surpass the pseudo label based approaches, e.g., ME outperforms both PLT and Co-training in \textbf{task3}. The underlying reason may be that when multiple organs are annotated in each partially labeled dataset, the mutual exclusiveness can be better used to regularize where the organ cannot overlap and thus reduce the ambiguity among different organs.

Second, we compare the approaches that exploit pseudo labels, including PLT\cite{liu2022universal}, Co-training\cite{huang2020multi} and MS-KD\cite{feng2021ms}. Compared to PLT where pseudo labels are not updated throughout the training process, Co-training uses a pair of co-trained networks to generate pseudo labels for each other and thus pseudo labels can be updated during training. Our results show that PLT and Co-training achieve comparable segmentation performance in \textbf{task1}. In \textbf{task2-4}, Co-training is among the top-performing methods and outperforms PLT consistently, suggesting that the quality of pseudo labels plays a key role for pseudo label learning. Besides, we observe unsatisfactory performance for the MS-KD, where the student model is trained solely on the soft pseudo labels generated by the teacher models. The underlying reason may be that the teacher models in MS-KD, i.e., the individual networks trained on each partially labeled dataset (Multi-Nets), are not strong. Hence, it may be necessary to incorporate both soft and hard labels (ground truth) as in \cite{hu2020knowledge} for more effective knowledge distillation. 


Third, we analyze the results achieved by the conditioned network, DoDNet\cite{zhang2021dodnet}. Overall, DoDNet achieves comparable segmentation performance to other methods in \textbf{task1} but sub-optimal performance in \textbf{task2-4}. For example, on the bowel datasets, it achieves inferior results on small structures such as duodenum and rectum compared to the methods that use multi-output channel networks. A possible reason could be that in our experiments we use the same backbone for DoDNet and other competing methods, but DoDNet may require a more complex backbone to achieve comparable results as in \cite{zhang2021dodnet} where the channels of decoder layers of DoDNet were doubled. Besides, we notice that DoDNet fails to distinguish the symmetric organs such as left and right femur/optic nerve, i.e., both sides of organs would be segmented when only asked for one side. The underlying reason may be that the conditioned networks by design learn each organ independently and thus may ignore the correlation among organs\cite{ye2023uniseg}. By contrast, multi-output channel networks, which segment all organs simultaneously, naturally capture the relationships among different organs. However, this suggests that DoDNet can be better at the segmentation tasks where organs are less correlated. For example, in \textbf{task3} (Tab. \ref{tab6}), we observe that DoDNet shows superior segmentation results on seminal vesicles and uterus, which are less correlated to other organs because they do not always appear due to gender difference. This finding aligns with the results presented in \cite{zhang2021dodnet} where DoDNet outperforms other methods in segmenting different types of tumors, which can be considered uncorrelated to each other. Lastly, since each organ is trained separately, DoDNet may be less efficient to train on the partially labeled dataset labeled with multiple organs. To summarize, the conditioned network DoDNet may need a more complex backbone to achieve optimal performance and is better at independent segmentation tasks.

\subsubsection{Analyses of COSST} 
The development of the proposed COSST is motivated by taking advantage of the effective components based on the empirical analyses above. Specifically, COSST is built upon a multi-output channel network by incorporating (1) mutual exclusiveness for regularization, (2) pseudo label for training, and (3) better pseudo labels for improved performance, where (1) and (2) correspond to the integration of comprehensive supervision signals and (3) corresponds to self-training and pseudo label filtering. In Tab. \ref{tab4}-\ref{tab7}, we show that the proposed COSST outperforms the top-performing benchmark methods, i.e., ME\cite{shi2021marginal} and Co-training\cite{huang2020multi}, on all four segmentation tasks with different degrees of improvement. Besides, in Fig. \ref{fig9}, we observe that ME outperforms Co-training by a large margin when the amount of training data is small, but slightly underperforms Co-training when more training data is available. Hence, given a new partial-label segmentation task, it is not clear which method in the literature should be adopted due to their sensitivity to the training data size. By contrast, COSST stands out as a more reliable option as it achieves consistent better performance than ME and Co-training regardless of the amount of training data. 

In Sec. \ref{sec:V}.D.1, we demonstrate the effectiveness of our pseudo label assessment approach with in-depth analyses. Specifically, we show that given the distribution of the ground truth labels, the quality of the unlabeled pseudo labels can be successfully assessed by using outlier detection in latent space. Our approach can thus be potentially extend to other fields where pseudo labeling is essential, such as semi-supervised learning and domain adaptation. However, this approach is far from perfect. In Fig. \ref{fig9}, we observe that rectum shows almost no correlation (PCC=-0.083) between the assessment metric and the actual Dice scores. This may be related to its widely dispersed distribution in latent space (Fig. \ref{fig8}) but further investigation is needed. Besides, our approach may be sensitive to the field of view (FOV) of images as the organ-wise feature representation is computed based on the organ mask. A reliable pseudo labels would be considered as outlier in latent space if its organ mask is not complete due to the limited FOV. In such scenario, we can selectively apply our approach to a subset of organs within the same FOV. 

\subsubsection{Future work}Learning from partially labeled datasets is critical to the emerging medical foundation models, which aim to train a universal segmentation model from large-scale datasets collected from different institutions. The types of supervision signals and the training strategy presented in our study can thus be used as a reference for future studies in foundation model development. For example, we observe that the current medical foundation models \cite{liu2023clip,ulrich2023multitalent,ye2023uniseg} have not exploited pseudo labels for training, i.e., Sup. III. Besides our study, modern foundation models in the computer vision community\cite{li2022grounded,kirillov2023segment} have also shown that self-training with pseudo labels is an effective technique for performance boost. Hence, it is a promising direction to incorporate pseudo label training to medical foundation models. Besides, we observe that nearly all existing studies for medical partial-label segmentation are focused on CT scans, possibly because CT scans collected from different institutes do not exhibit large domain gaps as in MRI. It is interesting for future studies to investigate partial-label segmentation in a cross-modality setting.

\noindent{\\}\textbf{Disclaimer}. The information in this paper is based on research results that are not commercially available. Future commercial availability cannot be guaranteed.

\bibliographystyle{IEEEtran}
\bibliography{manuscript}

\end{document}